%% file: main.tex
\documentclass{eitnlp}

\usepackage{wrapfig}            
\usepackage{colortbl}           
\usepackage{makecell}
\usepackage{multirow}
\usepackage{tikz}
\tcbuselibrary{most}

\definecolor{modelband}{HTML}{ECEFF7}
\definecolor{lvlEnc}{HTML}{F8E3DF}
\definecolor{lvlProj}{HTML}{FBEAD2}
\definecolor{projaccent}{HTML}{C07A1E}
\definecolor{lvlLLM}{HTML}{DDEBF7}
\definecolor{collapse}{HTML}{C0392B}
\definecolor{best}{HTML}{FFF6D5}
\definecolor{freegray}{HTML}{8C8C8C}
\definecolor{postblue}{HTML}{1A6FB0}
\definecolor{gain}{HTML}{C0392B}
\definecolor{drop}{HTML}{7F8C8D}
\definecolor{HeaderColor}{RGB}{245,247,250}
\definecolor{RatioColor}{RGB}{235,242,250}
\definecolor{groundtint}{HTML}{F5E6E0}

\DeclareRobustCommand{\circled}[1]{\tikz[baseline=(c.base)]{%
  \node[shape=circle,draw,inner sep=0.8pt,font=\tiny] (c) {#1};}}

\newcommand{\crit}[1]{#1\,{\textcolor{collapse}{\scriptsize$\blacktriangledown$}}}
\newcommand{\bgain}{\,\textcolor{gain}{\scriptsize$\blacktriangle$}}

\tcbset{
  findingbox/.style={
    colback=gray!3, colframe=gray!45!black, boxrule=0.5pt, arc=1mm,
    left=1.5mm, right=1.5mm, top=1mm, bottom=1mm,
    fontupper=\small, before skip=6pt, after skip=6pt
  }
}

\title{Rescuing Your Pruning: Rethinking Visual Token Pruning and Post-Training Recovery in MLLMs}
\runningtitle{Rescuing Your Pruning}
\author{\textbf{Peiran Yin}\textsuperscript{1,2,*}\qquad
        \textbf{Hao Wu}\textsuperscript{1,2,*}\qquad
        \textbf{Zihao Yang}\textsuperscript{3}\\
        \textbf{Haozhe Hu}\textsuperscript{1,2}\qquad
        \textbf{Xudong Wang}\textsuperscript{1}\qquad
        \textbf{Xiaoyu Shen}\textsuperscript{1,\textdagger}}
\affiliation{%
  \textsuperscript{1}EIT-NLP Lab, Eastern Institute of Technology, Ningbo\\
  \textsuperscript{2}Shanghai Jiao Tong University\qquad
  \textsuperscript{3}Uppsala University}
\authornotes{\textsuperscript{*}Equal contribution.\qquad
             \textsuperscript{\textdagger}Corresponding author.}
\contact{\texttt{peiran.yin@sjtu.edu.cn}\qquad\texttt{xyshen@eitech.edu.cn}}

\begin{document}
\maketitle

\begin{abstract}
Visual token pruning substantially reduces the inference cost of multimodal large language models (MLLMs), yet existing methods are mostly training-free and plug-and-play. They treat the resulting performance drop as an inevitable cost of compression and pursue increasingly complex pruning rules. We rethink this view through two questions: how much of the loss can post-training rescue, and, building on this recovery, how can joint compression push pruning to its limit? We curate a 282K-sample post-training mixture covering general VQA, reasoning, OCR, and grounding, and post-train 36 configurations that drop or merge tokens under a unified protocol. Post-training rescues most of the loss, even for OCR and grounding, and its adaptation gain in closing the gap between the pruned and the dense model grows as the budget tightens. For joint compression, we find that the vision encoder, not the LLM, dominates prefill time in native-resolution MLLMs, so where tokens are removed matters more than how many: keeping $40\%$ of tokens at the encoder runs faster than keeping $10\%$ at the projector. Our best configuration prunes the encoder lightly and the projector aggressively, leaves the LLM intact, and keeps general understanding and reasoning at the dense level. It accelerates the LLM stage by $3.49\times$ but the end-to-end time-to-first-token by only $1.54\times$, showing that LLM-stage speedups overstate end-to-end gains. We hope these findings offer insights for future research on joint visual token compression. Our data, protocol, and code will be released.
\end{abstract}

\input{Sec/1-introduction}
\input{Sec/2-related_work}   
\input{Sec/3-method}           
\input{Sec/4-experiments}      
\input{Sec/5-search}           
\input{Sec/5-conclusion}

\clearpage
\bibliographystyle{plainnat}
\bibliography{references}

\clearpage
\appendix
\input{Sec/appendix}

\end{document}

%% file: Sec/1-introduction.tex
\section{Introduction}
\label{sec:intro}

\begin{wrapfigure}{r}{0.48\textwidth}
    \centering
    \vspace{-12pt}
    \includegraphics[width=\linewidth]{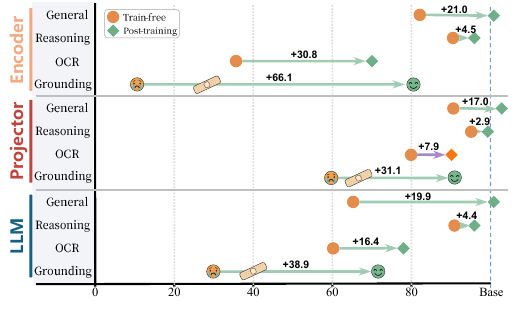}
    \caption{\textbf{Training-free vs.\ post-trained retention by capability.}
    Results on Qwen2.5-VL-7B at $20\%$ keep, with each pruning level averaged
    over its dropping and merging methods. Arrows mark the gain from
    post-training, and the dashed line marks the dense model.}
    \label{fig:motivation}
\end{wrapfigure}

Visual tokens, which often far outnumber text tokens, account for much of the inference cost of multimodal large language models (MLLMs), making visual token compression a primary route to acceleration~\citep{jin2025efficient, yao2026towards, wu2026hidrop, liu2026vica, tan2026vicostream}. Tokens can be reduced at three stages of the pipeline~\citep{Wu_2026}: inside the vision encoder~\citep{bolya2022token, jeddi2025similarity}, at the projector~\citep{alvar2025divprune, zhong2025aim, yang2025visionzip}, or within the LLM~\citep{chen2024image, xing2024pyramiddrop, dhouib2025pact}, by either dropping or merging them, mostly as training-free modules~\citep{zhang2024sparsevlm, shang2025llava, fan2025visipruner}. Rethinking this practice raises two questions. \textit{First, how much of the accuracy lost to pruning can be recovered?} As shown in Fig.~\ref{fig:motivation}, training-free pruning largely preserves coarse-grained understanding but sharply degrades OCR and grounding, consistent with prior observations~\citep{endo2025feather, chien2025grounding, huang2026n}. Yet lightweight post-training under the same pruning configuration pulls these nearly collapsed capabilities back to a large extent. Compression-aware post-training (CAPT) may thus mitigate the mismatch between the pruned input and a model trained on dense tokens~\citep{chien2025grounding, yang2025visionzip, shao2025growing}. \textit{Second, how should the three stages be compressed together?} A single stage offers limited room for compression, and combining stages can remove more tokens, as recent pipelines that chain several stages have begun to show~\citep{liu2025meteor, han2026filter, zhu2024focusllava}. Joint compression, however, couples the stages, and losses across stages accumulate. Early pruning shortens later sequences, but differing stage costs mean fewer tokens need not reduce end-to-end latency. By adapting the model under joint pruning, CAPT can mitigate the accumulated mismatch and keep more aggressive combinations feasible.

\begin{figure*}[t]
    \centering
    \includegraphics[width=1.0\textwidth]{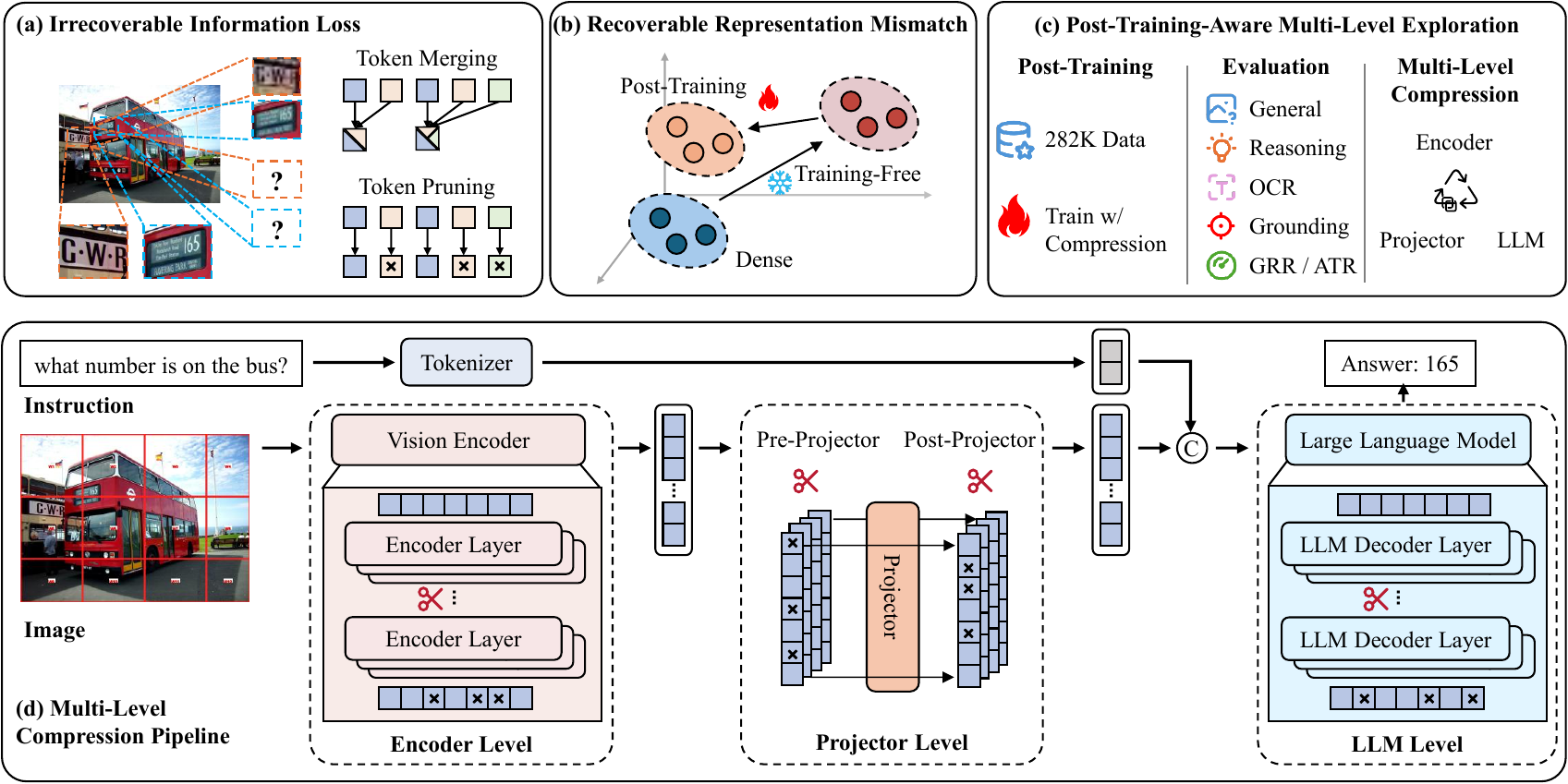}
    \caption{\textbf{Overview of our unified framework.}
    Pruning methods are organized by mechanism, dropping or merging, and by level, encoder, projector, or LLM. Post-training recovery is evaluated across General, Reasoning, OCR, and Grounding under single-stage and joint pruning configurations.}
    \label{fig:main_fig}
\end{figure*}

To study both questions systematically from the perspective of post-training, we first build a compact yet task-diverse training set of 282K samples, validate its composition through ablations, and check it for test contamination. Under a unified protocol that keeps every pruning rule unchanged, shown in Fig.~\ref{fig:main_fig}, we apply CAPT to 36 configurations: one dropping and one merging method at each stage, three keep ratios ($40\%$, $20\%$, and $10\%$), and two backbones, Qwen2.5-VL-7B~\citep{qwen2025qwen25vl} and LLaVA-OneVision-1.5-8B~\citep{an2025llavaonevision15}. We evaluate them on 12 benchmarks and report retention, the score relative to the unpruned model.

CAPT improves every configuration and recovers most of the performance lost to pruning. At an average keep ratio of $20\%$, it brings general understanding and reasoning close to the original model in nearly every configuration, and restores much of the OCR and grounding ability that training-free pruning nearly destroys. To trace where this recovery comes from, we take LLM-level pruning as a case study and contrast four training and evaluation states. They disentangle the pruning gap into a repairable adaptation mismatch and a residual of discarded evidence, and the post-training gain into a data gain and an adaptation gain. As the budget tightens, the residual grows, while the recovery relies increasingly on adapting to pruning. Visualizations further show that this adaptation acts on both sides of the pruning point. Upstream, the model learns features and attention that let the fixed rules select more relevant tokens. Downstream, the LLM learns to make better use of the few tokens that remain.

CAPT also opens room for joint compression. With CAPT, joint configurations that keep only about $5\%$ to $7\%$ of the visual tokens still preserve the original model's general understanding and reasoning, whereas recent training-free methods fall short of this level even at a $25\%$ budget. Exploring such configurations further leads to a more instructive finding: fewer tokens do not mean a faster model, and the configuration that keeps the fewest tokens is not the fastest. The question thus shifts from how many tokens to remove to where to remove them, and the answer depends on where the time goes. In native-resolution MLLMs such as Qwen2.5-VL, the encoder processes several times as many patches as the visual tokens the LLM receives after the projector's compression, and such encoders typically run less efficiently than their FLOP count suggests. \textit{Most of the prefill time therefore goes to the vision encoder rather than the 7B LLM}: $56.5\%$ in our setting, in line with the $40\%$ to over $80\%$ reported for similar models~\citep{zhu2026visionselector, liu2026pace, wang2026pixelprune, chen2025ipcv}. As a result, \textit{speedups measured inside the LLM overstate end-to-end gains}. Prior work often reports speedups of $3\times$ or more at little accuracy cost, measured on the LLM stage or on LLaVA-style models whose light encoders add little latency~\citep{chen2024image, zheng2025lossless, yang2025visionzip, wen2025stop}. On Qwen2.5-VL, keeping $10\%$ of the tokens at the projector speeds up the LLM stage by $3.38\times$, yet the end-to-end time-to-first-token (TTFT) by only $1.38\times$.\footnote{All methods are timed with one shared attention backend, PyTorch SDPA, as in prior work~\citep{jeddi2025similarity, zhang2025beyond}.} Real end-to-end speedups must therefore come from upstream, at the vision encoder and projector. Without training, however, upstream pruning sacrifices fine-grained capabilities. CAPT recovers them, making these extreme speedups feasible.

Under the same quality requirement, our search yields three guidelines for joint compression. (1) \textit{If only one stage can be compressed, it should be the encoder.} Keeping $40\%$ of the tokens at the encoder runs faster than keeping $10\%$ at the projector, because projector pruning leaves the encoder cost untouched and the LLM already approaches a length-independent fixed cost once about a hundred visual tokens remain (Appendix~\ref{app:llm_floor}). (2) \textit{If two stages can be compressed, light encoder pruning should be paired with aggressive projector pruning.} Our fastest feasible configuration, ToMe-70 with AIM-10, speeds up the LLM stage by $3.49\times$ and TTFT by $1.54\times$. (3) \textit{Compressing all three stages does not pay off.} LLM pruning adds operator overhead, and its accuracy loss forces the encoder to keep more tokens, so the three-stage combination is slower than the best two-stage one and even slower than encoder pruning alone. \textit{Where tokens are removed thus matters more than how many are kept}, which motivates accelerating the vision encoder itself in native-resolution MLLMs~\citep{chen2025ipcv, wang2026pixelprune, vasu2025fastvlm}.


Our contributions are as follows: (1) \textbf{Data and protocol.} We build a compact, task-diverse post-training mixture and a unified protocol that apply to dropping and merging methods at all three stages. (2) \textbf{Recoverability analysis.} A large-scale evaluation shows that post-training substantially improves pruned models. The pruning-induced degradation can be decomposed into information loss and distribution mismatch, and post-training partly repairs the mismatch, with this adaptation becoming more important as the budget tightens. (3) \textbf{Joint multi-stage compression.} CAPT with joint compression makes extreme visual token compression feasible, accelerating LLM prefill by $3.49\times$ and TTFT by $1.54\times$ without degrading general understanding and reasoning. End-to-end measurements further show that where tokens are removed matters more than how many are kept, and that LLM-stage speedups overstate end-to-end gains.

%% file: Sec/2-related_work.tex
\section{Related Work}
\label{sec:rel}




Visual token compression reduces the redundant visual sequence in MLLMs and has mainly developed along two directions: dropping and merging. Token dropping removes low-utility visual tokens by estimating token importance from attention patterns~\cite{chen2024image,xing2024pyramiddrop,yin2025lifting} or feature similarity~\cite{zhang2025beyond,alvar2025divprune}. FastV~\cite{chen2024image}, for example, prunes half of the visual tokens after the second LLM layer using the attention from the last text token to visual tokens. Token merging takes a softer route by fusing redundant token representations. AIM~\cite{zhong2025aim} directly merges similar token pairs, FiCoCo~\cite{han2026filter} recycles information by merging less important tokens into important retained tokens, and PACT~\cite{dhouib2025pact} clusters similar tokens into compact representatives. 
Unlike most training-free or plug-and-play studies, we revisit dropping and merging from a post-training perspective, analyzing how these two mechanisms behave after adaptation.

Efficient MLLM inference has motivated visual token compression beyond a single module, with recent methods combining reduction across multiple pipeline stages. FiCoCo~\cite{han2026filter} adopts an encoder--LLM pipeline with attention-guided pruning and information recycling, while FocusLLaVA~\cite{zhu2024focusllava} combines projector-level multi-scale sampling with LLM-level text-to-vision pruning. More recent methods extend compression to the encoder--projector--LLM pipeline: METEOR~\cite{liu2025meteor} integrates encoder pruning, cross-encoder redundancy removal, and instance-adaptive LLM pruning, and SAINT~\cite{jeddi2025similarity} applies similarity-aware pruning across stages. However, these methods mainly propose fixed multi-stage pipelines, leaving the interaction among stages and the allocation of compression budgets under the same final token budget insufficiently understood. We therefore systematically investigate joint encoder--projector--LLM compression and analyze how stage-wise budgets affect the accuracy--efficiency trade-off after adaptation.


%% file: Sec/3-method.tex
\section{Post-Training and Evaluation}
\label{sec:protocol}

\paragraph{Post-Training Data.}
Although some training-free pruning methods also report results with post-training, their data recipes always vary. We therefore first construct a unified, lightweight post-training dataset. Building upon the 133K LLaVA-Icons corpus~\cite{wu2024icons}, a compact subset of LLaVA-665K~\cite{liu2024improved}, we reorganize the originally multi-turn conversations into single-turn samples anchored by individual images. We then add small random portions of ChartQA~\cite{masry2022chartqa}, DVQA~\cite{kafle2018dvqa}, AI2D~\cite{kembhavi2016diagram}, DocVQA~\cite{mathew2021docvqa}, LLaVAR~\cite{zhang2023llavar}, and TQA~\cite{kembhavi2017you}, together with visual grounding data from LLaVA-665K, so that the mixture of about 282K samples covers general VQA, reasoning, OCR, and grounding. For visual grounding data used with Qwen-series models, we convert bounding boxes to normalized coordinates w.r.t.\ the original image resolution. Appendix~\ref{app:data_composition} details the dataset composition, Appendix~\ref{app:overlap} reports train--test overlap statistics, and Appendix~\ref{app:data_ablation} presents data ablations.

\paragraph{Training Protocol.}
Training-free pruning rules are designed for a frozen model, so we adapt them to CAPT under one uniform protocol. Pruning is a discrete operation defined by fixed arithmetic rather than learnable parameters, so we keep every rule exactly as specified and cut the gradient at the selection: what adapts is the model upstream and downstream of the pruning point, never the criterion itself. Pruning also reshapes the batch, since the number of surviving tokens varies across samples. We re-pack each sequence after every reduction, pad to the longest one in the batch, and extend the attention mask and labels so that padded positions enter neither attention nor the loss. For positions, retained tokens keep their pre-pruning indices and rotary embeddings are recomputed from those indices rather than the compacted ones, preserving the spatial layout of the dense model.

We experiment on Qwen2.5-VL-7B~\cite{qwen2025qwen25vl} and LLaVA-OneVision-1.5-8B (LLaVA-OV-1.5-8B)~\cite{an2025llavaonevision15}. LLM-level methods unfreeze only the language model. All other methods are trained with every module unfrozen. One setting deserves note: Qwen2.5-VL encodes images at dynamic resolution, so \texttt{max\_pixels} strongly affects fine-grained detail and training cost, and we fix it to $451{,}584$ as a balance between the two. The remaining configuration is shared across runs: one epoch over the corpus, a learning rate of $1\times10^{-5}$, bfloat16 precision, a maximum sequence length of 8192, and DeepSpeed ZeRO-2. Each run takes roughly four hours on 8 NVIDIA H20 80\,GB GPUs.

\paragraph{Evaluation Protocol and Metrics.}
We evaluate four capability groups with lmms-eval~\cite{zhang2024lmmsevalrealitycheckevaluation} under identical settings for all models. General ability is measured on GQA~\cite{hudson2019gqa}, VQA-v2~\cite{goyal2017making}, and OK-VQA~\cite{marino2019ok}, and reasoning on ScienceQA~\cite{lu2022learn}, AI2D~\cite{kembhavi2016diagram}, and POPE~\cite{li2023evaluating}, the last of which probes object-hallucination reasoning. OCR ability is measured on TextVQA~\cite{singh2019towards}, DocVQA~\cite{mathew2021docvqa}, and OCRBench~\cite{liu2024ocrbench}, and grounding on RefCOCO, RefCOCO+~\cite{kazemzadeh2014referitgame,yu2016modeling}, and RefCOCOg~\cite{mao2016generation}. We report ANLS for DocVQA and Acc@0.5 for grounding, where RefCOCO and RefCOCO+ are averaged over testA and testB and RefCOCOg uses its test split. All other benchmarks report accuracy.

To summarize performance under pruning, we report two retention metrics relative to the dense base model. The retention of a task is the pruned-model score divided by the dense-model score, expressed as a percentage. We first average retention within each group and then across groups. \emph{General-Reasoning Retention} (GRR) averages the General and Reasoning groups, and \emph{All-Task Retention} (ATR) averages all four groups.

\paragraph{Latency Measurement.}
All latency measurements use a single NVIDIA A100 GPU with batch size 1, SDPA attention, and the evaluation resolution, where \texttt{max\_pixels} is set to $1{,}605{,}632$. We report TTFT, measured with a single generated token, together with its breakdown into the vision encoder, the projector, and the LLM. Each number is averaged over $5{,}000$ TextVQA samples after five warm-up samples.

%% file: Sec/4-experiments.tex
\section{Rescuing Pruning Across Levels and Mechanisms}
\label{sec:exp}

To systematically investigate the effectiveness of post-training across pruning methods, we comprehensively explore three dimensions: the pruning mechanism (dropping vs.\ merging), the pruning position (encoder, projector, and LLM), and the compression strength ($40/20/10\%$). To account for dynamic-resolution inputs and reflect the actual computational budget, we use the average visual-token keep ratio per LLM layer as the invariant for all comparisons. Table~\ref{tab:enc_proj_free_post} summarizes the main results under encoder- and projector-level pruning, and Appendices~\ref{app:encoder}, \ref{app:projector}, and~\ref{app:llm} give the full results for each level.

\subsection{What Breaks and What Is Restored}
\label{sec:phenomena}

\begin{wrapfigure}{r}{0.48\textwidth}
    \centering
    \vspace{-35pt}
    \includegraphics[width=\linewidth]{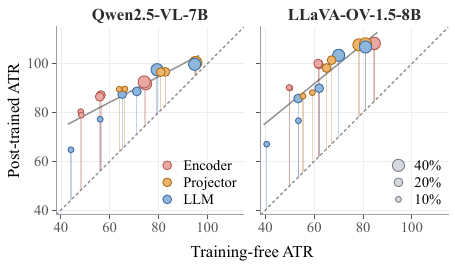}
    \vspace{-15pt}
    \caption{\textbf{Training-free vs.\ post-trained ATR} for 36 pruning
    configurations on two backbones. Color indicates the pruning level and
    marker size the keep ratio. The dashed line is $y=x$, vertical stems mark
    the post-training gain, and the gray line is a linear fit.}
    \label{fig:recovery_scatter}
    \vspace{-10pt}
\end{wrapfigure}
CAPT improves every configuration in the design space.
All $36$ configurations in Fig.~\ref{fig:recovery_scatter}, spanning two backbones, three pruning levels, two mechanisms, and three keep ratios, lie above the diagonal, so the recovery does not depend on a particular pruning setting.
The gain also grows with the severity of training-free degradation, and the fitted slopes stay below one, at $0.52$ on Qwen2.5-VL-7B and $0.81$ on LLaVA-OV-1.5-8B.
On Qwen2.5-VL-7B, for example, post-training improves ATR by only $5.3$ points for DivPrune at $40\%$ keep, which already starts close to the dense model, but by $32.2$ points for SAINT at $10\%$ keep.

\begin{table*}[!t]
\centering
\footnotesize
\setlength{\tabcolsep}{4pt}
\renewcommand{\arraystretch}{1.0}
\begin{tabular}{l c l | cccc @{\hspace{9pt}} cc | cccc @{\hspace{9pt}} cc}
\toprule
& & & \multicolumn{6}{c|}{\textbf{Qwen2.5-VL-7B}}
& \multicolumn{6}{c}{\textbf{LLaVA-OV-1.5-8B}} \\
\cmidrule(lr){4-7}\cmidrule(lr){8-9}\cmidrule(lr){10-13}\cmidrule(lr){14-15}
\textbf{Method} & \textbf{Keep} &
& Gen. & Reas. & OCR & Gr.\ & GRR & ATR
& Gen. & Reas. & OCR & Gr.\ & GRR & ATR \\
\midrule

\rowcolor{lvlEnc}\multicolumn{15}{l}{\textbf{Encoder-level}} \\
\multirow{6}{*}{SAINT}
& \multirow{2}{*}{$40\%$} & {\color{freegray}Free}
& 89.7 & 95.6 & 68.5 & 45.1 & 92.6 & 74.7
& 95.0 & 98.1 & 83.0 & 61.8 & 96.6 & 84.5 \\
& & {\color{postblue}Post}
& 104.1 & 97.3 & 80.7 & 84.6 & 100.7 & 91.7
& 100.0 & 98.5 & 90.2 & 144.2 & 99.2 & 108.2 \\
\addlinespace[1.5pt]
& \multirow{2}{*}{$20\%$} & {\color{freegray}Free}
& 81.7 & 91.6 & 39.6 & 12.5 & 86.7 & 56.4
& 86.4 & 94.0 & 46.4 & 20.7 & 90.2 & 61.9 \\
& & {\color{postblue}Post}
& 102.6 & 95.4 & 70.8 & 79.6 & 99.0 & 87.1
& 97.9 & 96.6 & 76.8 & 126.9 & 97.3 & 99.6 \\
\addlinespace[1.5pt]
& \multirow{2}{*}{$10\%$} & {\color{freegray}Free}
& 75.1 & 87.6 & 25.6 & \crit{4.0} & 81.4 & 48.1
& 76.6 & 88.4 & 29.9 & 5.2 & 82.5 & 50.0 \\
& & {\color{postblue}Post}
& 101.0 & 94.1 & 61.4 & 64.8\bgain & 97.5 & 80.3
& 97.3 & 93.6 & 64.6 & 104.7 & 95.4 & 90.0 \\
\cmidrule(l){1-15}
\multirow{6}{*}{ToMe}
& \multirow{2}{*}{$40\%$} & {\color{freegray}Free}
& 89.5 & 95.6 & 66.7 & 44.4 & 92.6 & 74.1
& 94.3 & 98.3 & 85.0 & 60.9 & 96.3 & 84.6 \\
& & {\color{postblue}Post}
& 104.9 & 97.5 & 80.2 & 87.5 & 101.2 & 92.5
& 100.8 & 97.9 & 89.7 & 144.5 & 99.3 & 108.3 \\
\addlinespace[1.5pt]
& \multirow{2}{*}{$20\%$} & {\color{freegray}Free}
& 81.7 & 91.2 & 36.3 & 14.6 & 86.4 & 55.9
& 86.6 & 94.0 & 44.6 & 21.2 & 90.3 & 61.6 \\
& & {\color{postblue}Post}
& 102.8 & 96.6 & 66.7 & 79.7 & 99.7 & 86.4
& 101.2 & 96.3 & 74.1 & 128.1 & 98.8 & 100.0 \\
\addlinespace[1.5pt]
& \multirow{2}{*}{$10\%$} & {\color{freegray}Free}
& 76.2 & 87.8 & 24.9 & \crit{4.5} & 82.0 & 48.4
& 76.8 & 88.4 & 29.5 & \crit{4.3} & 82.6 & 49.7 \\
& & {\color{postblue}Post}
& 101.0 & 94.0 & 56.4 & 64.3\bgain & 97.5 & 78.9
& 98.1 & 93.7 & 62.7 & 106.2\bgain & 95.9 & 90.2 \\
\midrule

\rowcolor{lvlProj}\multicolumn{15}{l}{\textbf{Projector-level}} \\
\multirow{6}{*}{DivPrune}
& \multirow{2}{*}{$40\%$} & {\color{freegray}Free}
& 96.0 & 98.8 & 94.0 & 91.5 & 97.4 & 95.1
& 95.9 & 98.3 & 89.6 & 29.4 & 97.1 & 78.3 \\
& & {\color{postblue}Post}
& 110.2 & 100.1 & 95.6 & 95.6 & 105.2 & 100.4
& 102.3 & 97.8 & 89.9 & 140.9 & 100.0 & 107.7 \\
\addlinespace[1.5pt]
& \multirow{2}{*}{$20\%$} & {\color{freegray}Free}
& 91.2 & 95.4 & 81.0 & 63.8 & 93.3 & 82.8
& 91.1 & 95.5 & 75.1 & 6.5 & 93.3 & 67.1 \\
& & {\color{postblue}Post}
& 108.5 & 98.3 & 89.3 & 90.3 & 103.4 & 96.6
& 102.4 & 96.3 & 80.6 & 126.3 & 99.4 & 101.4 \\
\addlinespace[1.5pt]
& \multirow{2}{*}{$10\%$} & {\color{freegray}Free}
& 83.4 & 90.5 & 63.8 & 27.2 & 87.0 & 66.2
& 84.0 & 91.2 & 60.2 & \crit{1.3} & 87.6 & 59.2 \\
& & {\color{postblue}Post}
& 105.3 & 95.6 & 79.1 & 78.5 & 100.4 & 89.6
& 99.1 & 93.4 & 69.9 & 89.9\bgain & 96.3 & 88.1 \\
\cmidrule(l){1-15}
\multirow{6}{*}{AIM}
& \multirow{2}{*}{$40\%$} & {\color{freegray}Free}
& 96.6 & 98.5 & 94.1 & 91.9 & 97.5 & 95.3
& 95.9 & 98.3 & 85.8 & 44.0 & 97.1 & 81.0 \\
& & {\color{postblue}Post}
& 110.3 & 99.6 & 95.6 & 96.1 & 104.9 & 100.4
& 104.8 & 97.9 & 87.5 & 142.3 & 101.3 & 108.1 \\
\addlinespace[1.5pt]
& \multirow{2}{*}{$20\%$} & {\color{freegray}Free}
& 92.6 & 95.7 & 79.8 & 55.4 & 94.1 & 80.9
& 90.9 & 93.7 & 61.1 & 14.3 & 92.3 & 65.0 \\
& & {\color{postblue}Post}
& 109.3 & 98.6 & 87.2 & 91.0 & 104.0 & 96.5
& 100.3 & 95.1 & 70.7 & 126.6 & 97.7 & 98.2 \\
\addlinespace[1.5pt]
& \multirow{2}{*}{$10\%$} & {\color{freegray}Free}
& 85.9 & 91.5 & 59.8 & 18.6 & 88.7 & 63.9
& 84.4 & 89.9 & 44.4 & \crit{3.0} & 87.2 & 55.4 \\
& & {\color{postblue}Post}
& 106.9 & 96.1 & 74.5 & 80.4 & 101.5 & 89.5
& 98.3 & 92.7 & 55.0 & 100.9\bgain & 95.5 & 86.7 \\
\bottomrule
\end{tabular}
\caption{\textbf{Per-task retention under encoder- and projector-level pruning,
{\color{freegray}before} and {\color{postblue}after} post-training.}
GRR averages General and Reasoning; ATR averages all four task groups.
\textcolor{collapse}{$\blacktriangledown$} flags training-free scores below $5$
(near-total failure); \textcolor{gain}{$\blacktriangle$} flags the paired
post-trained score, every one of which recovers above $60$.}
\label{tab:enc_proj_free_post}
\end{table*}

The impact of pruning varies considerably across tasks, as detailed for the encoder and projector levels in Table~\ref{tab:enc_proj_free_post}. Under training-free pruning, coarse-grained tasks remain relatively robust, and GRR never falls below $81$ at these two levels even at $10\%$ keep. In contrast, OCR degrades steadily as the token budget shrinks, and for SAINT on Qwen2.5-VL-7B its retention drops from $68.5$ at $40\%$ keep to $25.6$ at $10\%$. Grounding is even more fragile, collapsing to $4.0$ at $10\%$ keep on Qwen2.5-VL-7B and to single digits for every method on LLaVA-OV-1.5-8B. Post-training repairs these losses just as unevenly. For SAINT at $10\%$ keep on Qwen2.5-VL-7B, GRR rises from $81.4$ to $97.5$ and nearly reaches the dense level, while OCR rises from $25.6$ to $61.4$ and Grounding from $4.0$ to $64.8$. The fine-grained tasks thus gain far more but remain well below the dense model, and the GRR--ATR gap narrows from $33.3$ to $17.2$. The remaining gap widens as the token budget tightens, since post-trained ATR at $10\%$ keep stays below that at $20\%$ for every method on both backbones. Section~\ref{sec:attrib} decomposes this gap using LLM-level pruning as a case study.

The pruning mechanism reveals a different pattern. Dropping removes selected tokens entirely, whereas merging aggregates them into the surviving tokens and therefore preserves a compressed, albeit distorted, representation of the original visual evidence. Under training-free pruning, merging often shows a clear advantage, especially at aggressive compression. After CAPT, however, this advantage largely disappears at the encoder and projector levels, where matched dropping--merging pairs differ by at most $1.4$ ATR points on Qwen2.5-VL-7B and dropping can even perform slightly better, as SAINT reaches $80.3$ against $78.9$ for ToMe at $10\%$ keep. Inside the LLM, by contrast, PACT still outperforms PDrop by $12.5$ ATR points on Qwen2.5-VL-7B and by $9.6$ ATR points on LLaVA-OV-1.5-8B at $10\%$ keep. These results suggest that post-training can largely adapt to the representation distortion introduced by merging, causing much of its training-free advantage to vanish at earlier stages. When compression becomes more aggressive and is applied inside the LLM, however, preserving visual evidence remains important, whereas information discarded by dropping cannot be recovered through adaptation.

The overall recovery trend is consistent across the two backbones, although its magnitude is backbone-dependent. One caveat concerns grounding on LLaVA-OV-1.5-8B: after post-training, the dense model itself reaches $150.3$ retention relative to the Dense-Free baseline, indicating that the original model is substantially under-trained for grounding. Consequently, post-pruning grounding scores above $100$ on this backbone should not be interpreted as outperforming the corresponding adapted dense model.

\begin{figure*}[t]
    \centering
    \includegraphics[width=1.0\textwidth]{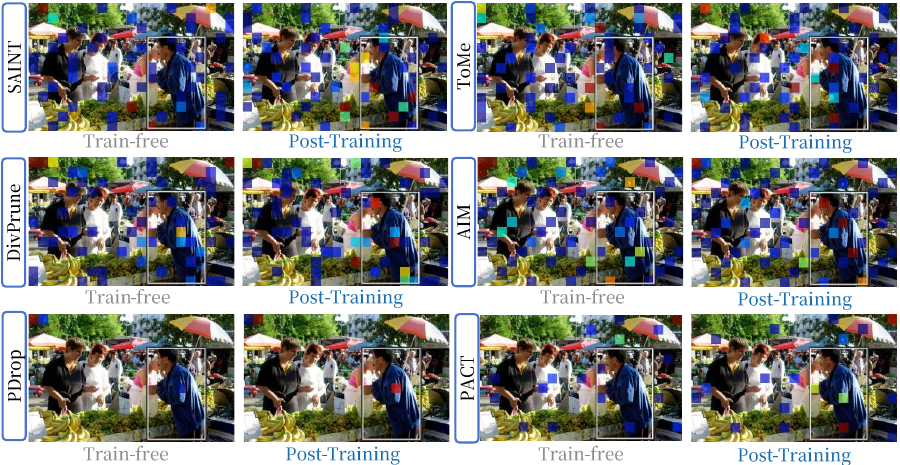}
    \caption{
    Retained visual tokens and layer-24 text-to-vision attention for six pruning
    methods at $20\%$ keep on a RefCOCO grounding example. Colored patches denote
    retained tokens, with color indicating relative attention from the last text
    token; transparent regions are pruned tokens, and the white box marks the
    ground-truth region.}
    \label{fig:attention_vis}
\end{figure*}

\subsection{Where Does the Recovery Come From?}
\label{sec:attrib}

The previous results show that post-training can recover a large fraction of the performance lost to visual-token pruning. Yet aggressively pruned models often remain below the dense reference even after training. This suggests a natural decomposition of the pruning-induced performance gap into two parts: a recoverable component that can be restored through post-training, and a residual component that mainly reflects the information loss inevitably introduced by pruning. We write

\begin{equation}
    \Delta_{\mathrm{gap}}
    =
    G_{\mathrm{recoverable}}
    +
    R_{\mathrm{res}}.
    \label{eq:gap_decomp}
\end{equation}

The recoverable component requires further attribution. Our post-training uses a small and diverse dataset to adapt the upstream and downstream layers of the pruning operation to the compressed token distribution, thereby mitigating pruning-induced mismatches such as the attention gap. However, introducing post-training data inevitably brings an additional source of performance gain that is independent of pruning adaptation. We therefore further decompose the recoverable gain into a \emph{data gain} and an \emph{adaptation gain}:

\begin{equation}
    G_{\mathrm{recoverable}}
    =
    G_{\mathrm{data}}
    +
    G_{\mathrm{adapt}}.
    \label{eq:recoverable_decomp}
\end{equation}

To quantify these components, we compare four experimental states for each
pruning rule and token budget:
\circled{1}~\textbf{Free}, the original model with pruning;
\circled{2}~\textbf{Data-only}, the dense post-trained model with pruning
enabled only at evaluation;
\circled{3}~\textbf{Prune-aware}, the model post-trained and evaluated with
pruning active, i.e., CAPT; and
\circled{4}~\textbf{Dense-PT}, the same model as \circled{2} evaluated without
pruning.

For either retention metric
$M\in\{\mathrm{GRR},\mathrm{ATR}\}$, let $M^{(i)}$ denote its value under
state \circled{$i$}. The three components are then given directly by
\begin{equation}
\begin{aligned}
G_{\mathrm{data}}
    &= M^{(2)} - M^{(1)}, \\
G_{\mathrm{adapt}}
    &= M^{(3)} - M^{(2)}, \\
R_{\mathrm{res}}
    &= M^{(4)} - M^{(3)}.
\end{aligned}
\label{eq:attribution}
\end{equation}
Thus, the recoverable gain and the full pruning gap are
\begin{equation}
\begin{aligned}
G_{\mathrm{recoverable}}
    &= M^{(3)} - M^{(1)}
     = G_{\mathrm{data}} + G_{\mathrm{adapt}}, \\
\Delta_{\mathrm{gap}}
    &= M^{(4)} - M^{(1)}
     = G_{\mathrm{recoverable}} + R_{\mathrm{res}}.
\end{aligned}
\label{eq:full_decomp}
\end{equation}
Finally, we report
\begin{equation}
\mathrm{RR}
=
\frac{G_{\mathrm{recoverable}}}{\Delta_{\mathrm{gap}}}
\times 100,
\label{eq:rr}
\end{equation}
which measures the fraction of the pruning-induced gap recovered through
post-training.

\begin{table*}[t]
\centering
\footnotesize
\setlength{\tabcolsep}{4.5pt}
\renewcommand{\arraystretch}{1.0}

\resizebox{\textwidth}{!}{%
\begin{tabular}{l c l | cccc | ccc | c | c}
\toprule
& & &
\multicolumn{4}{c|}{\textbf{Evaluation State}}
&
\multicolumn{3}{c|}{\textbf{Recovered Gain}}
&
\multirow{2}{*}{\textbf{Residual}}
&
\multirow{2}{*}{\textbf{RR}} \\

\cmidrule(lr){4-7}
\cmidrule(lr){8-10}

\textbf{Method}
& \textbf{Keep}
& \textbf{Metric}
& \circled{1} Free
& \circled{2} Data-only
& \circled{3} Prune-aware
& \circled{4} Dense-PT
& Total
& Data
& Adapt.
&
& \\

\midrule

\rowcolor{modelband}
\multicolumn{12}{c}{\textbf{Qwen2.5-VL-7B}} \\

\multirow{6}{*}{PDrop}
& \multirow{2}{*}{$40\%$} & GRR
& 92.5 & 101.0 & 103.9 & 106.0
& 11.4 & 8.5 & 2.9 & 2.1 & 84.4 \\
& & ATR
& 79.4 & 89.1 & 97.5 & 102.4
& 18.1 & 9.7 & 8.4 & 4.9 & 78.7 \\

\addlinespace[1.5pt]

& \multirow{2}{*}{$20\%$} & GRR
& 87.0 & 95.6 & 99.7 & 106.0
& 12.7 & 8.6 & 4.1 & 6.3 & 66.8 \\
& & ATR
& 65.1 & 72.5 & 87.5 & 102.4
& 22.4 & 7.4 & 15.0 & 14.9 & 60.1 \\

\addlinespace[1.5pt]

& \multirow{2}{*}{$10\%$} & GRR
& 66.8 & 76.4 & 89.6 & 106.0
& 22.8 & 9.6 & 13.2 & 16.4 & 58.2 \\
& & ATR
& 44.1 & 50.1 & 64.7 & 102.4
& 20.6 & 6.0 & 14.6 & 37.7 & 35.3 \\

\cmidrule(l){1-12}

\multirow{6}{*}{PACT}
& \multirow{2}{*}{$40\%$} & GRR
& 98.3 & 104.0 & 104.8 & 106.0
& 6.5 & 5.7 & 0.8 & 1.2 & 84.4 \\
& & ATR
& 94.8 & 96.3 & 99.7 & 102.4
& 4.9 & 1.5 & 3.4 & 2.7 & 64.5 \\

\addlinespace[1.5pt]

& \multirow{2}{*}{$20\%$} & GRR
& 90.0 & 97.8 & 101.5 & 106.0
& 11.5 & 7.8 & 3.7 & 4.5 & 71.9 \\
& & ATR
& 71.1 & 73.2 & 88.6 & 102.4
& 17.5 & 2.1 & 15.4 & 13.8 & 55.9 \\

\addlinespace[1.5pt]

& \multirow{2}{*}{$10\%$} & GRR
& 81.6 & 92.1 & 97.9 & 106.0
& 16.3 & 10.5 & 5.8 & 8.1 & 66.8 \\
& & ATR
& 56.1 & 62.0 & 77.2 & 102.4
& 21.1 & 5.9 & 15.2 & 25.2 & 45.6 \\

\midrule

\rowcolor{modelband}
\multicolumn{12}{c}{\textbf{LLaVA-OV-1.5-8B}} \\

\multirow{6}{*}{PDrop}
& \multirow{2}{*}{$40\%$} & GRR
& 90.2 & 93.1 & 98.4 & 101.1
& 8.2 & 2.9 & 5.3 & 2.7 & 75.5 \\
& & ATR
& 70.0 & 92.8 & 103.4 & 112.2
& 33.4 & 22.8 & 10.6 & 8.8 & 79.2 \\

\addlinespace[1.5pt]

& \multirow{2}{*}{$20\%$} & GRR
& 81.2 & 84.2 & 90.3 & 101.1
& 9.1 & 3.0 & 6.1 & 10.8 & 45.9 \\
& & ATR
& 53.4 & 69.4 & 85.7 & 112.2
& 32.3 & 16.0 & 16.3 & 26.5 & 54.9 \\

\addlinespace[1.5pt]

& \multirow{2}{*}{$10\%$} & GRR
& 68.3 & 74.0 & 85.5 & 101.1
& 17.2 & 5.7 & 11.6 & 15.6 & 52.6 \\
& & ATR
& 40.5 & 50.2 & 67.0 & 112.2
& 26.5 & 9.7 & 16.8 & 45.2 & 36.9 \\

\cmidrule(l){1-12}

\multirow{6}{*}{PACT}
& \multirow{2}{*}{$40\%$} & GRR
& 97.9 & 99.0 & 101.2 & 101.1
& 3.3 & 1.1 & 2.3 & -0.1 & 103.7 \\
& & ATR
& 81.0 & 102.7 & 106.8 & 112.2
& 25.8 & 21.7 & 4.1 & 5.5 & 82.5 \\

\addlinespace[1.5pt]

& \multirow{2}{*}{$20\%$} & GRR
& 91.2 & 92.6 & 97.5 & 101.1
& 6.3 & 1.4 & 4.8 & 3.6 & 63.3 \\
& & ATR
& 62.0 & 74.3 & 89.9 & 112.2
& 27.9 & 12.3 & 15.5 & 22.3 & 55.5 \\

\addlinespace[1.5pt]

& \multirow{2}{*}{$10\%$} & GRR
& 84.2 & 85.4 & 92.8 & 101.1
& 8.6 & 1.2 & 7.4 & 8.3 & 51.0 \\
& & ATR
& 53.5 & 59.3 & 76.6 & 112.2
& 23.1 & 5.8 & 17.2 & 35.7 & 39.3 \\

\bottomrule
\end{tabular}
}

\caption{\textbf{LLM-level recovery attribution.} All values are retention percentages relative to Dense-Free $=100$.}
\label{tab:llm_attrib}
\end{table*}

\paragraph{Aggressive pruning makes recovery more adaptation-dependent and more information-limited.}
Under aggressive budgets, $G_{\mathrm{adapt}}$ exceeds $G_{\mathrm{data}}$ in ATR and accounts for over $60\%$ of the recoverable gain at $10\%$ keep, showing that the benefit of CAPT does not come merely from the additional data. Once information loss becomes severe, however, adaptation reaches a ceiling. $G_{\mathrm{adapt}}$ stays nearly flat between $20\%$ and $10\%$ keep, while the residual keeps growing and more than doubles for PDrop on Qwen2.5-VL-7B. Since Prune-aware and Dense-PT use the same data and training recipe and differ only in whether pruning is applied, this uncompensated gap mainly stems from information loss and concentrates on fine-grained tasks.

\paragraph{CAPT rescues representation mismatch, but not discarded evidence.}
Dropping removes tokens outright, along with the information they carry. Merging instead aggregates this information into the retained tokens, at the cost of perturbing the original representations. Post-training adapts to this perturbation and lets the extra information kept by merging take effect. At $10\%$ keep, PACT and PDrop gain comparably from adaptation, yet PACT leaves a residual that is one fifth to one third smaller and outperforms PDrop by $12.5$ and $9.6$ points in ATR on the two backbones. Post-training thus determines how well the model exploits the compressed representation, while the information preserved by pruning sets the ceiling of recovery.



To qualitatively examine how post-training changes the pruned representation, we visualize the retained tokens and text-to-vision attention for all six methods at $20\%$ keep on a RefCOCO grounding example in Fig.~\ref{fig:attention_vis}. Upstream of the pruning point, post-training reshapes the feature and attention distributions that the pruning rules operate on, so that the rules select tokens more effectively and more retained visual tokens fall inside the ground-truth region. Downstream, the LLM assigns stronger relative attention to the retained tokens within the target region, indicating that it learns to exploit the remaining visual evidence more effectively. Together, these two effects constitute the adaptation gain quantified above.

%% file: Sec/5-search.tex
\section{How Far Can Post-Training Push Pruning?}
\label{sec:search}

Having established that post-training substantially improves the tolerance of MLLMs to aggressive visual-token pruning, we ask how much prefill speedup can be achieved while general understanding and reasoning remain essentially lossless. Under the latency setting of Section~\ref{sec:protocol}, dense Qwen2.5-VL-7B takes $213.06$\,ms TTFT per sample, of which the vision encoder accounts for $120.43$\,ms, the projector for $0.65$\,ms, and the LLM for $85.10$\,ms. The projector itself is almost free, yet it is a natural compression point, because pruning there shortens the visual sequence processed by the entire LLM. Projector-level pruning also preserves quality at aggressive budgets. As shown in Section~\ref{sec:phenomena}, at $10\%$ keep on Qwen2.5-VL-7B, post-trained encoder-level ToMe retains only $78.9$ ATR, whereas projector-level AIM retains $89.5$. We therefore fix AIM at $10\%$ keep, denoted AIM-10, as a projector anchor and search how much additional compression the encoder and the LLM can absorb on top of it.

\begin{figure*}[ht]
    \centering
    \includegraphics[width=\textwidth]{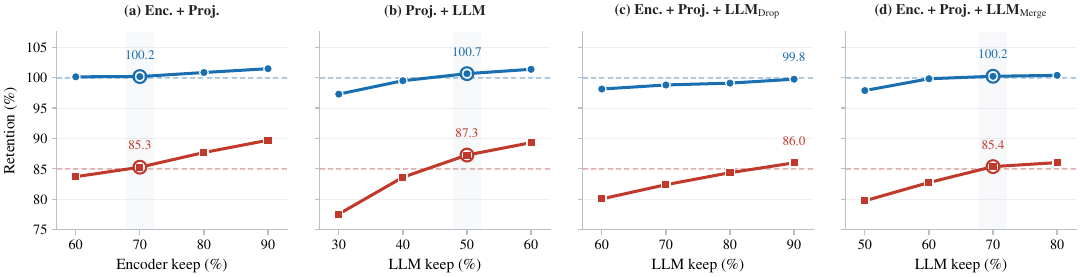}
    \caption{\textbf{Multi-level compression search with post-training.}
    AIM is fixed at $10\%$ keep at the projector. (a) Encoder-level ToMe with a
    dense LLM. (b) LLM-level PACT with a dense encoder. (c) LLM-level PDrop and
    (d) LLM-level PACT with ToMe fixed at $80\%$. Blue circles and red squares
    denote GRR and ATR, dashed lines mark the constraints $\mathrm{GRR}=100$ and
    $\mathrm{ATR}=85$, and highlighted points mark the lowest feasible keep ratio.}
    \label{fig:search_ablation}
\end{figure*}


At the encoder we use ToMe, which matches SAINT in latency but performs better, as compared in Appendix~\ref{app:encoder_operator}. To locate the compression limit of each level under essentially lossless performance, we search over keep ratios subject to a quality constraint. We require GRR to stay at or above $100$, i.e., general understanding and reasoning at roughly the dense level, and ATR to stay at or above $85$, and we select the lowest keep ratio that satisfies both, as shown in Fig.~\ref{fig:search_ablation}. With additional compression at a single level, the encoder can be pushed to ToMe-70 and the LLM to PACT-50. To also reduce the encoder, which dominates latency, we further combine all three levels. Since losses accumulate across levels, each level must retain more tokens when combined, and the feasible three-level configuration becomes ToMe-80, AIM-10, and PACT-70. Because the clustering and merging of PACT introduce extra overhead, we also try replacing it with the lighter PDrop. Under the same combination, however, PDrop fails to satisfy the constraint even at $90\%$ keep, indicating that the information discarded by dropping becomes increasingly intolerable near the compression limit and that preserving visual evidence matters more.

CAPT remains effective for such multi-level joint pruning, keeping these aggressive combinations within the feasible region. Fig.~\ref{fig:search_ablation}, however, only locates this frontier in token space. Pruning at different levels removes different amounts of computation and introduces different operator costs. Since computational savings do not necessarily translate into proportional speedups~\citep{hu2026beyond}, a smaller token budget need not yield lower latency. We therefore measure their actual prefill latency in Table~\ref{tab:search_prefill}.

\begin{table}[t]
\centering
\small
\setlength{\tabcolsep}{7pt}
\setlength{\aboverulesep}{0pt}
\setlength{\belowrulesep}{0pt}
\renewcommand{\arraystretch}{1.22}
\newcommand{\grp}[1]{\rowcolor{gray!8}\multicolumn{6}{@{\hspace{6pt}}l}{\textit{#1}}}
\newcommand{\spd}[1]{\,{\scriptsize(#1$\times$)}}
\begin{tabular}{@{\hspace{6pt}}l cc ccc@{\hspace{6pt}}}
\toprule
& \multicolumn{2}{c}{\textbf{Quality}}
& \multicolumn{3}{c}{\textbf{Latency in ms (speedup)}} \\
\cmidrule(lr){2-3}\cmidrule(lr){4-6}
\textbf{Method}
& \textbf{GRR} & \textbf{ATR}
& \textbf{E+P} & \textbf{L} & \textbf{TTFT} \\
\midrule
Dense & 100.0 & 100.0 & 121.08\spd{1.00} & 85.10\spd{1.00} & 213.06\spd{1.00} \\
\midrule
\grp{Recent training-free methods, 25\% budget} \\
DART~\citep{wen2025stop} & 88.9 & 68.1 & 121.00\spd{1.00} & 35.51\spd{2.40} & 162.05\spd{1.31} \\
EADP~\citep{eadp} & 91.8 & 82.1 & 151.88\spd{0.80} & 30.44\spd{2.80} & 186.74\spd{1.14} \\
VLM-Pruner~\citep{vlmpruner} & 93.6 & 80.1 & 118.89\spd{1.02} & 51.99\spd{1.64} & 174.73\spd{1.22} \\
VisionZip~\citep{yang2025visionzip}  & 86.9 & 80.6 & 129.46\spd{0.94} & 41.04\spd{2.07} & 175.63\spd{1.21} \\
ToMe~\citep{bolya2022token}       & 85.7 & 57.6 &  86.43\spd{1.40} & 30.32\spd{2.81} & 129.91\spd{1.64} \\
AIM~\citep{zhong2025aim}        & 93.6 & 84.9 & 123.79\spd{0.98} & 30.12\spd{2.83} & 159.53\spd{1.34} \\
PDrop~\citep{xing2024pyramiddrop}      & 82.5 & 63.7 & 120.88\spd{1.00} & 37.28\spd{2.28} & 164.08\spd{1.30} \\
PACT~\citep{dhouib2025pact}       & 92.5 & 81.6 & 121.19\spd{1.00} & 44.35\spd{1.92} & 171.54\spd{1.24} \\
\grp{Compression-aware post-training} \\
P10         & 101.5 & 89.5 & 124.50\spd{0.97} & 25.15\spd{3.38} & 154.54\spd{1.38} \\
\rowcolor{postblue!10}
\textbf{E70+P10} & \textbf{100.2} & \textbf{85.3} & \textbf{108.91\spd{1.11}} & \textbf{24.38\spd{3.49}} & \textbf{138.61\spd{1.54}} \\
E80+P10+L70 & 100.2 & 85.4 & 114.28\spd{1.06} & 32.71\spd{2.60} & 152.51\spd{1.40} \\
P10+L50     & 100.7 & 87.3 & 122.04\spd{0.99} & 31.66\spd{2.69} & 158.67\spd{1.34} \\
\bottomrule
\end{tabular}
\caption{\textbf{Quality and prefill latency of our feasible configurations and of training-free pruning methods.}
E, P, and L denote pruning at the encoder, projector, and LLM, each followed by its keep ratio in percent. Each latency is followed by its speedup over the Dense row; the speedup of TTFT is the end-to-end speedup.}
\label{tab:search_prefill}
\end{table}


On the one hand, post-training enables acceleration while satisfying the
quality constraints. E70+P10 reduces TTFT from $213.06$ to $138.61$\,ms, a $1.54\times$
end-to-end speedup, while keeping GRR at the dense level. Recent training-free
methods fall short in quality, and all but ToMe also in speed. Even with $25\%$ of the visual
tokens retained, none of them reaches dense-level GRR, and apart from ToMe
their end-to-end speedups range only from $1.14\times$ to $1.34\times$, suggesting that
training-free pruning is approaching a plateau.

On the other hand, the stage breakdown shows that MLLM acceleration cannot be
judged by the LLM stage alone. AIM-10 speeds up the LLM stage by $3.38\times$,
but the vision side stays almost unchanged and takes about four fifths of
TTFT, so the end-to-end speedup is only $1.38\times$. With encoder pruning
added, E70+P10 speeds up the vision side by $1.11\times$ alongside a
$3.49\times$ faster LLM stage, lifting the end-to-end speedup to $1.54\times$
and making it the fastest feasible configuration. Training-free ToMe, which
prunes upstream at the encoder, likewise has by far the smallest E+P and the
highest overall speedup in the table at $1.64\times$, albeit with a severe
quality loss. The AIM-10 experiments also reveal a counterintuitive effect.
Once tokens are aggressively pruned before the LLM, further LLM-level pruning
leaves latency nearly unchanged or even slightly higher, because the LLM
already operates close to a length-independent latency floor, as analyzed in
Appendix~\ref{app:llm_floor}. Joint pruning across levels is therefore
worthwhile for real acceleration, and CAPT is what rescues the
quality of such aggressive joint configurations.

%% file: Sec/5-conclusion.tex
\section{Conclusion}
\label{sec:conclusion}

This paper revisits visual token compression in multimodal large language models from the perspective of post-training. We point out that the performance degradation after pruning stems not only from unrecoverable information loss, but also from the adaptation gap between compressed visual tokens and the original dense-token distribution. To this end, we establish a unified compression-aware post-training framework and systematically analyze the recovery behavior of different compression mechanisms at the encoder, projector, and LLM levels. Our framework achieves significant and comprehensive performance recovery across diverse pruning settings, suggesting that a substantial portion of the degradation caused by training-free pruning is adaptable. Furthermore, we explore aggressive joint pruning across levels, which, with compression-aware post-training, attains the highest speedup among nearly lossless configurations.

%% file: Sec/appendix.tex


\section{Post-Training Data Details}
\label{app:data}

\subsection{Dataset Composition}
\label{app:data_composition}

Table~\ref{tab:posttrain_data} summarizes the data sources in our post-training mixture. The bulk comes from the single-turn reorganization of LLaVA-Icons~\citep{wu2024icons}, together with the visual grounding data of LLaVA-665K~\citep{liu2024improved}, which consists of region annotations on VG and COCO images. We further add targeted supplements for OCR and reasoning, and retain a small language-only split from ShareGPT to preserve text-only instruction following. The final mixture totals about 282K samples.

\begin{table}[htbp]
\centering
\small
\renewcommand{\arraystretch}{1.18}
\setlength{\tabcolsep}{10pt}
\begin{tabular}{l r l}
\toprule
\textbf{Dataset} & \textbf{Count} & \textbf{Category} \\
\midrule
COCO~\citep{lin2014microsoft}     & 103{,}909 & General \\
VG~\citep{krishna2017visual}       &  55{,}504 & Grounding \\
GQA~\citep{hudson2019gqa}      &  42{,}777 & Reasoning \\
OCRVQA~\citep{mishra2019ocr}   &  27{,}346 & General + OCR \\
ChartQA~\citep{masry2022chartqa}  &  10{,}000 & Reasoning + OCR \\
DVQA~\citep{kafle2018dvqa}     &  10{,}000 & Reasoning + OCR \\
DocVQA~\citep{mathew2021docvqa}   &  10{,}000 & OCR \\
AI2D~\citep{kembhavi2016diagram}     &   7{,}508 & Reasoning \\
LLaVAR~\citep{zhang2023llavar}   &   6{,}785 & OCR \\
TQA~\citep{kembhavi2017you}      &   6{,}501 & Reasoning \\
ShareGPT~\citep{chiang2023vicuna} &   1{,}000 & Language-only \\
TextVQA~\citep{singh2019towards}  &      963  & General + OCR \\
\midrule
\textbf{Total} & \textbf{282{,}293} & -- \\
\bottomrule
\end{tabular}
\caption{Composition of the post-training data mixture, listed by data source.}
\label{tab:posttrain_data}
\end{table}

\subsection{Train--Test Overlap}
\label{app:overlap}

To check for train--test contamination, we compare the six supplementary datasets added beyond LLaVA-Icons and LLaVA-665K, namely ChartQA, DVQA, AI2D, DocVQA, LLaVAR, and TQA, against all evaluation benchmarks. As shown in Table~\ref{tab:data_decontamination}, only 8 AI2D images and 2 DocVQA images also appear in the supplementary data, and none of them is paired with the same question, so the supplementary data contains no evaluation sample.

\begin{table}[htbp]
\centering
\small
\renewcommand{\arraystretch}{1.15}
\setlength{\tabcolsep}{8pt}
\begin{tabular}{l rr rrr}
\toprule
\textbf{Benchmark} & \textbf{Samples} & \textbf{Images} & \textbf{Img.} & \textbf{Q.} & \textbf{Triplet} \\
\midrule
GQA       & 12{,}578  &    398   & 0 & 0 & 0 \\
VQAv2     & 214{,}354 & 40{,}504 & 0 & 0 & 0 \\
OK-VQA    & 5{,}046   & 5{,}033  & 0 & 0 & 0 \\
\midrule
ScienceQA & 2{,}017   & 1{,}238  & 0 & 0 & 0 \\
AI2D      & 3{,}088   &    814   & 8 & 0 & 0 \\
POPE      & 9{,}000   &    500   & 0 & 0 & 0 \\
\midrule
TextVQA   & 5{,}000   & 3{,}166  & 0 & 0 & 0 \\
DocVQA    & 5{,}349   & 1{,}284  & 2 & 0 & 0 \\
OCRBench  & 1{,}000   &    930   & 0 & 0 & 0 \\
\midrule
RefCOCO   & 10{,}752  & 1{,}500  & 0 & 0 & 0 \\
RefCOCO+  & 10{,}615  & 1{,}500  & 0 & 0 & 0 \\
RefCOCOg  & 9{,}602   & 2{,}600  & 0 & 0 & 0 \\
\bottomrule
\end{tabular}
\caption{Overlap between the supplementary post-training data and the evaluation benchmarks. Img., Q., and Triplet count the evaluation images, image--question pairs, and image--question--answer triplets that also appear in the supplementary data, respectively.}
\label{tab:data_decontamination}
\end{table}

\subsection{Data Ablation}
\label{app:data_ablation}

To verify the diversity and soundness of our post-training mixture, we run four ablation groups on Qwen2.5-VL-7B with projector-level AIM at $10\%$ keep (Table~\ref{tab:posttrain_data_ablation}). The first group, \emph{w/o Direct Eval}, removes GQA, AI2D, TextVQA, and DocVQA, the training sources that share their origin with an evaluation benchmark, to test whether the gains come from source overlap. Each of the other three removes one task category: \emph{w/o OCR\,/\,Text-rich} drops OCR-oriented data, \emph{w/o Grounding} drops all samples with bounding-box annotations, and \emph{w/o Chart\,/\,Diagram} drops chart and diagram data. Every removal is paired with a \emph{Random} control that drops the same number of samples uniformly at random, separating the effect of \emph{which} data is removed from \emph{how much}.

\begin{table*}[t]
\centering
\scriptsize
\renewcommand{\arraystretch}{1.28}
\setlength{\tabcolsep}{3.6pt}
\resizebox{\textwidth}{!}{%
\begin{tabular}{l ccc ccc ccc ccc cc}
\toprule
& \multicolumn{3}{c}{\textbf{General}} & \multicolumn{3}{c}{\textbf{Reasoning}} & \multicolumn{3}{c}{\textbf{OCR}} & \multicolumn{3}{c}{\textbf{Grounding}} & \multicolumn{2}{c}{\textbf{Aggregate}} \\
\cmidrule(lr){2-4} \cmidrule(lr){5-7} \cmidrule(lr){8-10} \cmidrule(lr){11-13} \cmidrule(lr){14-15}
\textbf{Training Data} & \textbf{GQA} & \textbf{VQA-v2} & \textbf{OK-VQA} & \textbf{SciQA} & \textbf{AI2D} & \textbf{POPE} & \textbf{TextVQA} & \textbf{DocVQA} & \textbf{OCRBench} & \textbf{RefCOCO} & \textbf{RefCOCO+} & \textbf{RefCOCOg} & \textbf{GRR} & \textbf{ATR} \\
\midrule
Training-Free & 56.4 & 74.8 & 32.8 & 83.2 & 70.5 & 82.9 & 60.7 & 48.2 & 46.1 & 17.6 & 15.2 & 14.6 & 88.7 & 63.9 \\
\rowcolor{blue!6}
Full Mixture & 60.7 & 77.5 & 55.9 & 83.5 & 76.9 & 88.1 & 70.9 & 66.4 & 56.8 & 73.1 & 64.3 & 67.6 & 101.5 & 89.5 \\
\midrule
w/o Direct Eval & 57.2 & 77.4 & 56.3 & 83.1 & 75.8 & 87.5 & 70.4 & 62.0 & 54.9 & 72.7 & 63.7 & 66.4 & 100.3 & 88.0 \\
\quad Random & 59.9 & 77.5 & 56.4 & 82.5 & 76.6 & 87.2 & 70.7 & 65.1 & 57.2 & 70.1 & 60.7 & 64.0 & 101.1 & 88.2 \\
\midrule
w/o OCR / Text-rich & 60.7 & 77.6 & 56.1 & 83.8 & 76.6 & 87.5 & 70.1 & 60.2 & 54.0 & 72.4 & 63.8 & 67.3 & 101.5 & 88.5 \\
\quad Random & 60.4 & 77.3 & 56.9 & 82.4 & 76.5 & 87.3 & 71.2 & 65.4 & 55.3 & 70.2 & 60.9 & 65.3 & 101.4 & 88.4 \\
\midrule
w/o Grounding & 60.1 & 77.6 & 58.2 & 83.1 & 76.2 & 87.4 & 71.0 & 65.7 & 57.4 & 8.7 & 7.3 & 6.6 & 101.9 & 71.8 \\
\quad Random & 59.8 & 77.3 & 55.7 & 82.1 & 76.1 & 87.5 & 71.0 & 65.2 & 57.2 & 70.9 & 62.3 & 65.1 & 100.6 & 88.4 \\
\midrule
w/o Chart / Diagram & 60.5 & 77.4 & 57.6 & 82.8 & 75.2 & 86.9 & 70.9 & 64.0 & 54.9 & 71.2 & 62.3 & 65.5 & 101.4 & 88.5 \\
\quad Random & 61.2 & 77.4 & 56.7 & 83.5 & 75.7 & 88.3 & 70.7 & 65.6 & 57.0 & 70.7 & 61.2 & 64.6 & 101.7 & 88.7 \\
\bottomrule
\end{tabular}}
\caption{\small \textbf{Data ablation on Qwen2.5-VL-7B with AIM at $10\%$ keep at the projector.}
All runs use the full post-training recipe of the main experiments (all modules trainable, one epoch). Training-Free is AIM-10 without post-training.
Each ``w/o'' row removes one group of training sources; the paired ``Random'' row removes the same number of samples uniformly at random.
The \colorbox{blue!6}{highlighted} row is the full mixture. POPE reports accuracy; RefCOCO and RefCOCO+ average testA and testB, and RefCOCOg uses test.}
\label{tab:posttrain_data_ablation}
\vspace{-8pt}
\end{table*}

The results support our data design. First, the full mixture achieves the highest ATR and every removal lowers it, while GRR stays above the dense level in all post-trained runs, indicating that general and reasoning abilities are carried by the bulk of the data and the targeted supplements mainly serve fine-grained abilities. Second, the gains do not stem from source overlap. Removing all Direct-Eval sources leaves ATR on par with its random control, and GQA falls noticeably below the random control, while AI2D, TextVQA, and DocVQA remain well above the training-free level. Third, each targeted removal hurts the ability it supplies more than a random removal of the same size, most visibly on DocVQA without OCR data. Fourth, grounding data is irreplaceable. Without it, Acc@0.5 on the RefCOCO family falls below $10\%$, even lower than before post-training, and ATR drops to $71.8$.

\section{Cross-Ratio Generalization}
\label{app:cross_ratio}
In deployment, the keep ratio may differ from the one used in post-training. To test whether a post-trained model transfers across budgets, we post-train LLM-level PACT on Qwen2.5-VL-7B at $10\%$, $20\%$, and $40\%$ keep and evaluate each model at all three ratios (Table~\ref{tab:cross_ratio_generalization}). Matching the training and inference ratios gives the best ATR in every column, so post-training is most effective when it targets the deployed budget. The transfer is asymmetric, however. The model trained at $10\%$ keep loses only $3.9$ ATR points when evaluated at $40\%$, whereas the model trained at $40\%$ loses $6.2$ points when evaluated at $10\%$. Adapting to a tight budget thus carries over to looser budgets, but not the other way around. GRR varies by less than $4$ points within each column, so the mismatch again mainly affects fine-grained tasks.

\begin{table}[ht]
\centering
\small
\setlength{\tabcolsep}{4pt}
\renewcommand{\arraystretch}{1.08}
\begin{tabular}{cc ccc}
\toprule
\textbf{Train Keep} & \textbf{Metric} & \multicolumn{3}{c}{\textbf{Inference Keep Ratio (\%)}} \\
\cmidrule(lr){3-5}
\textbf{Ratio} & & \textbf{10} & \textbf{20} & \textbf{40} \\
\midrule
\multirow{2}{*}{10\%} & GRR & 97.9 & 101.2 & 104.1 \\
& ATR & 77.2 & 86.8 & 95.8 \\
\addlinespace[2pt]
\multirow{2}{*}{20\%} & GRR & 96.6 & 101.5 & 104.4 \\
& ATR & 76.0 & 88.6 & 98.7 \\
\addlinespace[2pt]
\multirow{2}{*}{40\%} & GRR & 94.0 & 99.4 & 104.8 \\
& ATR & 71.0 & 84.8 & 99.7 \\
\bottomrule
\end{tabular}
\caption{\small Cross-ratio generalization of LLM-level PACT on Qwen2.5-VL-7B. Rows give the keep ratio used in post-training and columns the keep ratio at inference. Diagonal entries use matched training and inference ratios.}
\label{tab:cross_ratio_generalization}
\vspace{-8pt}
\end{table}

\section{Details of the Multi-Level Search}
\label{app:search_details}
This section provides two supporting analyses for the multi-level search in Section~\ref{sec:search}: the choice of the encoder operator and the latency floor of the LLM.

\subsection{Encoder Operator Selection}
\label{app:encoder_operator}

Table~\ref{tab:encoder_operator} compares the two encoder-level operators at $40\%$ keep on Qwen2.5-VL-7B. SAINT and ToMe reach virtually identical prefill latency, while ToMe attains slightly higher GRR and ATR after post-training. Since the multi-level search in Section~\ref{sec:search} constrains GRR, we adopt ToMe as the encoder operator.

\begin{table}[ht]
\centering
\footnotesize
\renewcommand{\arraystretch}{1.08}
\setlength{\tabcolsep}{10pt}
\begin{tabular}{l ccc}
\toprule
\textbf{Method} & \textbf{TTFT (ms)} & \textbf{GRR} & \textbf{ATR} \\
\midrule
SAINT & \textbf{143.99} & 100.7 & 91.7 \\
ToMe  & 144.11 & \textbf{101.2} & \textbf{92.5} \\
\bottomrule
\end{tabular}
\caption{Encoder-level operators at $40\%$ keep on Qwen2.5-VL-7B, measured under the latency setting of Section~\ref{sec:protocol} (dense: $213.06$\,ms).}
\label{tab:encoder_operator}
\end{table}

\subsection{LLM Latency Floor}
\label{app:llm_floor}

\begin{wrapfigure}{r}{0.45\textwidth}
\centering
\vspace{-12pt}
\includegraphics[width=\linewidth]{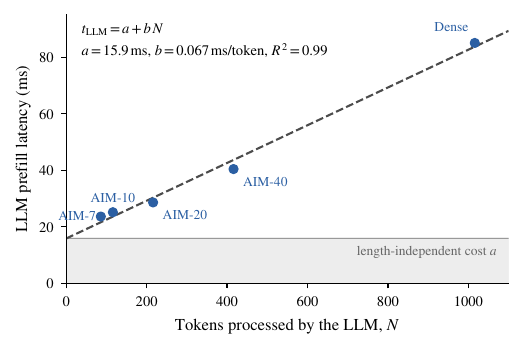}
\caption{LLM prefill latency versus the number of tokens processed by the LLM on Qwen2.5-VL-7B. Points are dense and projector-level AIM at $40/20/10/7\%$ keep, and the dashed line is the affine fit. The shaded band marks the length-independent cost $a$.}
\label{fig:llm_floor}
\vspace{-10pt}
\end{wrapfigure}
Table~\ref{tab:search_prefill} shows that adding LLM-level pruning on top of AIM-10 does not reduce latency. To explain this, we vary only the number of tokens $N$ entering the LLM, using dense Qwen2.5-VL-7B and projector-level AIM at $40$, $20$, $10$, and $7\%$ keep, and fit the LLM prefill latency as $t_{\mathrm{LLM}} = a + bN$, as shown in Fig.~\ref{fig:llm_floor}. The fit gives $a \approx 15.9$\,ms and $b \approx 0.067$\,ms per token. Under AIM-10, the LLM processes about $116$ tokens, so only about $8$\,ms of its $25.15$\,ms latency still depends on sequence length. Further LLM-level pruning can save at most this amount, yet it adds an operator cost of its own. As a result, adding PACT-50 on top of AIM-10 raises TTFT from $154.54$ to $158.67$\,ms. Reducing encoder computation is not bounded by this floor, which is why E70+P10 is the fastest feasible configuration even though it retains more tokens than E80+P10+L70 and P10+L50. Once the LLM approaches its floor, token count is no longer a reliable proxy for latency.

\section{Details of Pruning Methods}
\label{app:methods}

We study one dropping and one merging method at each pruning level: SAINT and ToMe at the vision encoder, DivPrune and AIM at the projector, and PDrop and PACT inside the LLM.

\paragraph{SAINT~\cite{jeddi2025similarity}.}
SAINT can be applied in a training-free manner to prune visual tokens by estimating token redundancy from key-embedding similarity graphs and using layer-adaptive pruning rates across the multimodal pipeline.

\paragraph{ToMe~\cite{bolya2022token}.}
ToMe is a training-free token merging method that accelerates Vision Transformers by iteratively matching similar tokens and merging them in pairs inside transformer blocks, thereby reducing token redundancy without additional training.

\paragraph{DivPrune~\cite{alvar2025divprune}.}
DivPrune reduces visual tokens by formulating pruning as a Max-Min Diversity Problem, retaining a diverse subset of tokens while removing visually redundant or less informative ones.

\paragraph{AIM~\cite{zhong2025aim}.}
AIM improves multimodal inference efficiency by first merging similar visual tokens before the LLM and then progressively pruning less important visual tokens inside LLM layers.

\paragraph{PyramidDrop (PDrop)~\cite{xing2024pyramiddrop}.}
PyramidDrop progressively reduces visual tokens across transformer layers with a pyramid-like schedule, removing more tokens in earlier stages while preserving more tokens in later layers.

\paragraph{PACT~\cite{dhouib2025pact}.}
PACT reduces visual tokens by pruning less informative tokens and clustering nearby redundant tokens into representative embeddings.

\section{Details of Datasets and Benchmarks}
\label{app:benchmarks}

\paragraph{COCO~\cite{lin2014microsoft}.}
COCO is a large-scale dataset of natural images with rich object, segmentation, and caption annotations, widely used for image captioning and general visual understanding.

\paragraph{VG~\cite{krishna2017visual}.}
Visual Genome (VG) provides dense annotations of objects, attributes, and relationships grounded in image regions, supporting fine-grained scene understanding and region-level visual grounding.

\paragraph{OCRVQA~\cite{mishra2019ocr}.}
OCR-VQA consists of book cover images paired with questions about titles, authors, and other textual content, requiring joint OCR and visual question answering.

\paragraph{ChartQA~\cite{masry2022chartqa}.}
ChartQA evaluates question answering over chart images, requiring both visual perception of chart elements and logical or arithmetic reasoning over the underlying data.

\paragraph{DVQA~\cite{kafle2018dvqa}.}
DVQA evaluates understanding of bar charts through questions about structure, data retrieval, and reasoning, with answers often containing chart-specific vocabulary.

\paragraph{LLaVAR~\cite{zhang2023llavar}.}
LLaVAR augments visual instruction tuning with text-rich images, enhancing the ability of multimodal models to read and reason over textual content embedded in images.

\paragraph{TQA~\cite{kembhavi2017you}.}
TQA (Textbook Question Answering) evaluates multimodal reasoning over middle-school science textbooks, combining text passages, diagrams, and questions that require contextual understanding.

\paragraph{ShareGPT~\cite{chiang2023vicuna}.}
ShareGPT is a collection of user-shared conversations with ChatGPT, commonly used as a language-only instruction-following corpus to preserve linguistic capability during multimodal fine-tuning.

\paragraph{GQA~\cite{hudson2019gqa}.}
GQA evaluates compositional visual reasoning based on scene-graph annotations from Visual Genome.

\paragraph{VQA-v2~\cite{goyal2017making}.}
VQA-v2 is a balanced open-ended visual question answering benchmark designed to reduce language priors and encourage visual grounding.

\paragraph{OK-VQA~\cite{marino2019ok}.}
OK-VQA evaluates knowledge-based visual question answering, where answering often requires external or commonsense knowledge beyond the image content.

\paragraph{ScienceQA~\cite{lu2022learn}.}
ScienceQA evaluates multimodal scientific reasoning over questions from natural, language, and social sciences, where the image-based subset requires models to reason over visual inputs.

\paragraph{AI2D~\cite{kembhavi2016diagram}.}
AI2D evaluates diagram understanding, requiring models to interpret labels, symbols, visual structures, and spatial relations.

\paragraph{POPE~\cite{li2023evaluating}.}
POPE is a diagnostic benchmark for object hallucination, using binary questions to test whether models falsely report nonexistent objects.

\paragraph{TextVQA~\cite{singh2019towards}.}
TextVQA evaluates scene-text understanding by requiring models to recognize and reason over text in natural images.

\paragraph{DocVQA~\cite{mathew2021docvqa}.}
DocVQA evaluates document image question answering, emphasizing OCR, layout understanding, and information localization.

\paragraph{OCRBench~\cite{liu2024ocrbench}.}
OCRBench evaluates OCR-related capabilities of MLLMs across text recognition, document understanding, information extraction, and related tasks.

\paragraph{RefCOCO~\cite{kazemzadeh2014referitgame,yu2016modeling}.}
RefCOCO evaluates referring expression comprehension by requiring models to localize objects described by natural-language expressions.

\paragraph{RefCOCO+~\cite{kazemzadeh2014referitgame,yu2016modeling}.}
RefCOCO+ emphasizes appearance-based referring expressions with fewer absolute location cues, making it more dependent on fine-grained object attributes.

\paragraph{RefCOCOg~\cite{mao2016generation,yu2016modeling}.}
RefCOCOg contains longer and more descriptive referring expressions, testing fine-grained and context-rich visual grounding.

\section{Extended Experimental Results}
\label{app:per-benchmark}

\subsection{Encoder-Level}
\label{app:encoder}
This section provides the full results of encoder-level pruning across different models, pruning ratios, and compression mechanisms. Overall, SAINT and ToMe achieve comprehensive improvements across metrics after post-training, with especially clear recovery on fine-grained abilities such as OCR and grounding. These results further suggest that encoder-level pruning contains a large adaptable degradation component, although extreme compression still leaves an information-loss upper bound.
\begin{table*}[p]
\centering
\scriptsize
\renewcommand{\arraystretch}{1.28}
\setlength{\tabcolsep}{2.1pt}

\resizebox{\textwidth}{!}{
\begin{tabular}{@{}l l c l *{12}{c} c c@{}}
\toprule
\multirow{2}{*}{\textbf{Backbone}} & \multirow{2}{*}{\textbf{Method}} & \multirow{2}{*}{\textbf{Ratio}} & \multirow{2}{*}{\textbf{Setting}} & \multicolumn{3}{c}{\textbf{General}} & \multicolumn{3}{c}{\textbf{Reasoning}} & \multicolumn{3}{c}{\textbf{OCR}} & \multicolumn{3}{c}{\textbf{Grounding}} & \multirow{2}{*}{\textbf{GRR}} & \multirow{2}{*}{\textbf{ATR}} \\
\cmidrule(lr){5-7}
\cmidrule(lr){8-10}
\cmidrule(lr){11-13}
\cmidrule(lr){14-16}
& & & & GQA & VQA$^{\mathrm{v2}}$ & OK & SQA & AI2D & POPE & TVQA & VQA$^{\mathrm{Doc}}$ & OCRB & Ref & Ref+ & RefG & (\%) & (\%) \\
\midrule
\multirow{13}{*}{\textbf{Qwen2.5-VL}} & \textbf{Base} & - & - & \textbf{60.9} & \textbf{82.9} & \textbf{43.8} & \textbf{88.2} & \textbf{82.5} & \textbf{87.6} & \textbf{82.8} & \textbf{94.7} & \textbf{83.6} & \textbf{88.2} & \textbf{82.2} & \textbf{84.4} & \textbf{100.0} & \textbf{100.0} \\
\addlinespace[1pt]
& \multirow{6}{*}{SAINT} & \multirow{2}{*}{40} & {\color{freegray}Free} & 57.5 & 76.9 & 35.9 & 84.7 & 77.8 & 84.4 & 62.8 & 71.7 & 45.2 & 39.9 & 36.9 & 38.2 & 92.6 & 74.7 \\
& & & {\color{postblue}Post} & 60.5 & 77.3 & 52.4 & 84.7 & 80.3 & 86.2 & 69.0 & 84.7 & 57.9 & 77.3 & 66.5 & 72.0 & 100.7 & 91.7 \\
& & \multirow{2}{*}{20} & {\color{freegray}Free} & 55.5 & 72.5 & 29.1 & 81.3 & 72.9 & 82.7 & 44.8 & 30.0 & 27.7 & 11.2 & 10.6 & 10.1 & 86.7 & 56.4 \\
& & & {\color{postblue}Post} & 60.2 & 75.1 & 51.8 & 82.3 & 78.6 & 85.4 & 61.1 & 75.9 & 48.9 & 72.6 & 63.3 & 67.1 & 99.0 & 87.1 \\
& & \multirow{2}{*}{10} & {\color{freegray}Free} & 52.5 & 68.4 & 24.8 & 78.0 & 68.7 & 79.9 & 30.9 & 16.2 & 18.8 & 4.0 & 3.7 & 2.6 & 81.4 & 48.1 \\
& & & {\color{postblue}Post} & 58.4 & 73.4 & 51.9 & 81.6 & 76.5 & 84.9 & 58.0 & 61.4 & 41.3 & 60.2 & 50.3 & 54.7 & 97.5 & 80.3 \\
\addlinespace[2pt]
& \multirow{6}{*}{ToMe} & \multirow{2}{*}{40} & {\color{freegray}Free} & 58.0 & 76.5 & 35.5 & 83.9 & 78.5 & 84.6 & 58.7 & 70.8 & 45.5 & 39.6 & 36.2 & 37.4 & 92.6 & 74.1 \\
& & & {\color{postblue}Post} & 61.2 & 77.3 & 53.0 & 84.8 & 80.4 & 86.6 & 67.9 & 86.3 & 56.5 & 78.7 & 70.2 & 74.2 & 101.2 & 92.5 \\
& & \multirow{2}{*}{20} & {\color{freegray}Free} & 55.1 & 72.8 & 29.2 & 81.3 & 71.0 & 83.6 & 42.2 & 26.0 & 25.5 & 13.4 & 12.4 & 11.4 & 86.4 & 55.9 \\
& & & {\color{postblue}Post} & 60.2 & 75.7 & 51.8 & 83.7 & 79.0 & 86.8 & 60.2 & 71.9 & 43.0 & 72.4 & 63.4 & 67.3 & 99.7 & 86.4 \\
& & \multirow{2}{*}{10} & {\color{freegray}Free} & 52.4 & 69.3 & 25.8 & 78.5 & 67.7 & 81.0 & 30.6 & 14.8 & 18.5 & 4.4 & 4.1 & 3.1 & 82.0 & 48.4 \\
& & & {\color{postblue}Post} & 58.5 & 73.9 & 51.6 & 81.6 & 75.5 & 85.8 & 55.2 & 55.5 & 36.8 & 60.4 & 49.8 & 53.9 & 97.5 & 78.9 \\
\midrule
\multirow{13}{*}{\textbf{LLaVA-OV-1.5}} & \textbf{Base} & - & - & \textbf{62.2} & \textbf{81.9} & \textbf{48.7} & \textbf{98.5} & \textbf{84.4} & \textbf{89.2} & \textbf{79.6} & \textbf{97.0} & \textbf{81.7} & \textbf{56.3} & \textbf{50.5} & \textbf{51.1} & \textbf{100.0} & \textbf{100.0} \\
\addlinespace[1pt]
& \multirow{6}{*}{SAINT} & \multirow{2}{*}{40} & {\color{freegray}Free} & 61.2 & 79.4 & 43.7 & 96.7 & 81.7 & 88.6 & 72.0 & 84.5 & 58.3 & 34.9 & 30.6 & 32.1 & 96.6 & 84.5 \\
& & & {\color{postblue}Post} & 60.5 & 79.1 & 51.7 & 96.9 & 82.7 & 88.3 & 71.7 & 90.6 & 71.3 & 79.6 & 72.2 & 75.7 & 99.2 & 108.2 \\
& & \multirow{2}{*}{20} & {\color{freegray}Free} & 58.9 & 74.8 & 35.7 & 91.6 & 77.1 & 87.1 & 52.4 & 31.5 & 33.5 & 12.1 & 9.8 & 10.9 & 90.2 & 61.9 \\
& & & {\color{postblue}Post} & 60.1 & 77.4 & 50.0 & 94.9 & 80.4 & 87.7 & 66.2 & 76.7 & 55.8 & 71.3 & 62.3 & 66.8 & 97.3 & 99.6 \\
& & \multirow{2}{*}{10} & {\color{freegray}Free} & 55.8 & 68.6 & 27.4 & 85.2 & 71.2 & 84.1 & 35.8 & 17.3 & 22.0 & 3.0 & 2.5 & 2.7 & 82.5 & 50.0 \\
& & & {\color{postblue}Post} & 58.0 & 75.2 & 52.0 & 90.6 & 77.5 & 86.4 & 59.6 & 61.1 & 45.6 & 60.0 & 50.6 & 54.8 & 95.4 & 90.0 \\
\addlinespace[2pt]
& \multirow{6}{*}{ToMe} & \multirow{2}{*}{40} & {\color{freegray}Free} & 61.0 & 79.5 & 42.8 & 96.6 & 82.4 & 88.5 & 71.6 & 84.3 & 63.8 & 35.0 & 29.7 & 31.5 & 96.3 & 84.6 \\
& & & {\color{postblue}Post} & 61.0 & 79.2 & 52.4 & 96.9 & 81.5 & 88.0 & 71.8 & 90.4 & 70.0 & 79.8 & 72.0 & 76.2 & 99.3 & 108.3 \\
& & \multirow{2}{*}{20} & {\color{freegray}Free} & 59.1 & 75.2 & 35.5 & 91.5 & 77.2 & 87.0 & 51.3 & 29.7 & 31.6 & 12.6 & 9.8 & 11.2 & 90.3 & 61.6 \\
& & & {\color{postblue}Post} & 60.1 & 77.8 & 54.6 & 94.1 & 80.2 & 87.8 & 64.9 & 73.2 & 53.4 & 71.5 & 63.5 & 67.2 & 98.8 & 100.0 \\
& & \multirow{2}{*}{10} & {\color{freegray}Free} & 55.7 & 69.4 & 27.3 & 84.7 & 71.3 & 84.5 & 36.7 & 16.9 & 20.3 & 2.7 & 1.9 & 2.2 & 82.6 & 49.7 \\
& & & {\color{postblue}Post} & 58.1 & 75.8 & 52.8 & 90.7 & 77.4 & 86.8 & 60.4 & 57.7 & 43.2 & 60.6 & 51.6 & 55.5 & 95.9 & 90.2 \\
\bottomrule
\end{tabular}
}
\caption{
Full encoder-level pruning results on Qwen2.5-VL-7B and LLaVA-OV-1.5-8B.
Ratio is the layer-averaged visual-token keep ratio in percent, and Free and Post denote the model before and after post-training.
VQA$^{\mathrm{v2}}$ and VQA$^{\mathrm{Doc}}$ denote VQA-v2 and DocVQA.
}
\label{tab:saint_tome_free_post}
\end{table*}


\subsection{Projector-Level}
\label{app:projector}
This section reports the complete experimental results of projector-level pruning. DivPrune and AIM show stable and comprehensive improvements after post-training. Under $40\%$ and $20\%$ keep ratios, their overall performance approaches or even surpasses the Dense-Free baseline. These results indicate that projector-level pruning is a stable single-level pruning choice and provides a strong basis for multi-level combinations.
\begin{table*}[p]
\centering
\scriptsize
\renewcommand{\arraystretch}{1.28}
\setlength{\tabcolsep}{2.1pt}

\resizebox{\textwidth}{!}{
\begin{tabular}{@{}l l c l *{12}{c} c c@{}}
\toprule
\multirow{2}{*}{\textbf{Backbone}} & \multirow{2}{*}{\textbf{Method}} & \multirow{2}{*}{\textbf{Ratio}} & \multirow{2}{*}{\textbf{Setting}} & \multicolumn{3}{c}{\textbf{General}} & \multicolumn{3}{c}{\textbf{Reasoning}} & \multicolumn{3}{c}{\textbf{OCR}} & \multicolumn{3}{c}{\textbf{Grounding}} & \multirow{2}{*}{\textbf{GRR}} & \multirow{2}{*}{\textbf{ATR}} \\
\cmidrule(lr){5-7}
\cmidrule(lr){8-10}
\cmidrule(lr){11-13}
\cmidrule(lr){14-16}
& & & & GQA & VQA$^{\mathrm{v2}}$ & OK & SQA & AI2D & POPE & TVQA & VQA$^{\mathrm{Doc}}$ & OCRB & Ref & Ref+ & RefG & (\%) & (\%) \\
\midrule
\multirow{13}{*}{\textbf{Qwen2.5-VL}} & \textbf{Base} & - & - & \textbf{60.9} & \textbf{82.9} & \textbf{43.8} & \textbf{88.2} & \textbf{82.5} & \textbf{87.6} & \textbf{82.8} & \textbf{94.7} & \textbf{83.6} & \textbf{88.2} & \textbf{82.2} & \textbf{84.4} & \textbf{100.0} & \textbf{100.0} \\
\addlinespace[1pt]
& \multirow{6}{*}{DivPrune} & \multirow{2}{*}{40} & {\color{freegray}Free} & 60.0 & 80.9 & 40.2 & 86.8 & 81.5 & 86.8 & 79.8 & 88.0 & 77.6 & 81.2 & 74.9 & 77.1 & 97.4 & 95.1 \\
& & & {\color{postblue}Post} & 63.0 & 80.7 & 56.9 & 87.8 & 82.1 & 88.6 & 80.1 & 91.2 & 78.4 & 85.1 & 77.3 & 81.4 & 105.2 & 100.4 \\
& & \multirow{2}{*}{20} & {\color{freegray}Free} & 58.5 & 78.1 & 36.5 & 83.9 & 77.2 & 85.3 & 73.4 & 73.2 & 64.4 & 58.1 & 51.8 & 52.6 & 93.3 & 82.8 \\
& & & {\color{postblue}Post} & 61.6 & 79.4 & 56.2 & 85.0 & 81.2 & 87.6 & 78.3 & 84.5 & 70.4 & 80.8 & 72.5 & 76.8 & 103.4 & 96.6 \\
& & \multirow{2}{*}{10} & {\color{freegray}Free} & 55.7 & 73.8 & 30.6 & 80.5 & 70.9 & 82.5 & 62.6 & 54.6 & 48.7 & 25.3 & 21.9 & 22.1 & 87.0 & 66.2 \\
& & & {\color{postblue}Post} & 59.8 & 76.9 & 54.7 & 82.2 & 78.0 & 86.6 & 73.8 & 72.7 & 59.6 & 71.4 & 62.2 & 66.5 & 100.4 & 89.6 \\
\addlinespace[2pt]
& \multirow{6}{*}{AIM} & \multirow{2}{*}{40} & {\color{freegray}Free} & 60.1 & 81.2 & 40.8 & 86.8 & 81.2 & 86.5 & 80.5 & 88.4 & 76.8 & 81.8 & 75.8 & 76.6 & 97.5 & 95.3 \\
& & & {\color{postblue}Post} & 63.4 & 81.1 & 56.5 & 86.7 & 82.2 & 88.4 & 80.2 & 90.9 & 78.5 & 85.8 & 77.1 & 82.1 & 104.9 & 100.4 \\
& & \multirow{2}{*}{20} & {\color{freegray}Free} & 58.7 & 78.8 & 37.8 & 85.2 & 76.9 & 85.1 & 73.9 & 71.8 & 62.3 & 51.1 & 45.4 & 44.7 & 94.1 & 80.9 \\
& & & {\color{postblue}Post} & 62.5 & 79.8 & 56.5 & 86.0 & 80.3 & 88.5 & 77.3 & 81.7 & 68.7 & 81.8 & 73.3 & 76.8 & 104.0 & 96.5 \\
& & \multirow{2}{*}{10} & {\color{freegray}Free} & 56.4 & 74.8 & 32.8 & 83.2 & 70.5 & 82.9 & 60.7 & 48.2 & 46.1 & 17.6 & 15.2 & 14.6 & 88.7 & 63.9 \\
& & & {\color{postblue}Post} & 60.7 & 77.5 & 55.9 & 83.5 & 76.9 & 88.1 & 70.9 & 66.4 & 56.8 & 73.1 & 64.3 & 67.6 & 101.5 & 89.5 \\
\midrule
\multirow{13}{*}{\textbf{LLaVA-OV-1.5}} & \textbf{Base} & - & - & \textbf{62.2} & \textbf{81.9} & \textbf{48.7} & \textbf{98.5} & \textbf{84.4} & \textbf{89.2} & \textbf{79.6} & \textbf{97.0} & \textbf{81.7} & \textbf{56.3} & \textbf{50.5} & \textbf{51.1} & \textbf{100.0} & \textbf{100.0} \\
\addlinespace[1pt]
& \multirow{6}{*}{DivPrune} & \multirow{2}{*}{40} & {\color{freegray}Free} & 60.9 & 79.7 & 45.0 & 95.9 & 82.3 & 89.2 & 77.5 & 84.5 & 68.8 & 16.9 & 14.2 & 15.2 & 97.1 & 78.3 \\
& & & {\color{postblue}Post} & 60.9 & 80.3 & 54.0 & 95.7 & 81.9 & 88.3 & 75.5 & 85.8 & 70.5 & 77.7 & 70.0 & 74.4 & 100.0 & 107.7 \\
& & \multirow{2}{*}{20} & {\color{freegray}Free} & 59.7 & 77.3 & 40.4 & 92.6 & 79.1 & 88.2 & 73.2 & 65.5 & 53.8 & 3.9 & 3.0 & 3.5 & 93.3 & 67.1 \\
& & & {\color{postblue}Post} & 60.2 & 79.0 & 55.5 & 93.8 & 80.3 & 87.9 & 73.4 & 72.4 & 61.1 & 70.1 & 62.2 & 66.9 & 99.4 & 101.4 \\
& & \multirow{2}{*}{10} & {\color{freegray}Free} & 56.8 & 73.6 & 34.4 & 87.5 & 74.6 & 85.9 & 65.7 & 45.8 & 41.5 & 0.7 & 0.7 & 0.7 & 87.6 & 59.2 \\
& & & {\color{postblue}Post} & 58.4 & 76.7 & 53.4 & 90.1 & 77.2 & 86.6 & 69.2 & 57.6 & 51.7 & 53.0 & 41.5 & 47.6 & 96.3 & 88.1 \\
\addlinespace[2pt]
& \multirow{6}{*}{AIM} & \multirow{2}{*}{40} & {\color{freegray}Free} & 61.5 & 80.0 & 44.4 & 96.6 & 82.0 & 88.8 & 74.8 & 81.3 & 65.0 & 25.2 & 21.0 & 23.4 & 97.1 & 81.0 \\
& & & {\color{postblue}Post} & 61.8 & 80.3 & 57.0 & 96.8 & 81.0 & 88.5 & 74.7 & 83.9 & 67.1 & 78.1 & 71.1 & 75.2 & 101.3 & 108.1 \\
& & \multirow{2}{*}{20} & {\color{freegray}Free} & 60.3 & 77.3 & 39.6 & 91.4 & 75.5 & 88.1 & 63.5 & 46.5 & 45.3 & 8.5 & 6.5 & 7.6 & 92.3 & 65.0 \\
& & & {\color{postblue}Post} & 60.5 & 79.1 & 52.2 & 93.3 & 77.5 & 88.1 & 67.6 & 60.1 & 53.3 & 70.6 & 62.1 & 67.0 & 97.7 & 98.2 \\
& & \multirow{2}{*}{10} & {\color{freegray}Free} & 58.3 & 73.4 & 34.0 & 87.5 & 71.1 & 86.2 & 51.9 & 28.4 & 31.6 & 2.0 & 1.4 & 1.4 & 87.2 & 55.4 \\
& & & {\color{postblue}Post} & 58.8 & 76.7 & 51.9 & 90.6 & 73.7 & 87.9 & 58.6 & 40.3 & 40.8 & 58.7 & 48.3 & 52.5 & 95.5 & 86.7 \\
\bottomrule
\end{tabular}
}
\caption{
Full projector-level pruning results on Qwen2.5-VL-7B and LLaVA-OV-1.5-8B.
Ratio is the layer-averaged visual-token keep ratio in percent, and Free and Post denote the model before and after post-training.
}
\label{tab:divprune_aim_free_post}
\end{table*}

\subsection{LLM-Level}
\label{app:llm}
This section presents the full results of LLM-level pruning. The results show that both PDrop and PACT obtain overall improvements on general, reasoning, OCR, and grounding metrics after post-training, with PACT usually showing more stable performance. This demonstrates that LLM-side pruning can also be effectively adapted through post-training, although fine-grained abilities remain more vulnerable under extreme compression.
\begin{table*}[p]
\centering
\scriptsize
\renewcommand{\arraystretch}{1.28}
\setlength{\tabcolsep}{2.1pt}

\resizebox{\textwidth}{!}{
\begin{tabular}{@{}l l c l *{12}{c} c c@{}}
\toprule
\multirow{2}{*}{\textbf{Backbone}} & \multirow{2}{*}{\textbf{Method}} & \multirow{2}{*}{\textbf{Ratio}} & \multirow{2}{*}{\textbf{Setting}} & \multicolumn{3}{c}{\textbf{General}} & \multicolumn{3}{c}{\textbf{Reasoning}} & \multicolumn{3}{c}{\textbf{OCR}} & \multicolumn{3}{c}{\textbf{Grounding}} & \multirow{2}{*}{\textbf{GRR}} & \multirow{2}{*}{\textbf{ATR}} \\
\cmidrule(lr){5-7}
\cmidrule(lr){8-10}
\cmidrule(lr){11-13}
\cmidrule(lr){14-16}
& & & & GQA & VQA$^{\mathrm{v2}}$ & OK & SQA & AI2D & POPE & TVQA & VQA$^{\mathrm{Doc}}$ & OCRB & Ref & Ref+ & RefG & (\%) & (\%) \\
\midrule
\multirow{13}{*}{\textbf{Qwen2.5-VL}} & \textbf{Base} & - & - & \textbf{60.9} & \textbf{82.9} & \textbf{43.8} & \textbf{88.2} & \textbf{82.5} & \textbf{87.6} & \textbf{82.8} & \textbf{94.7} & \textbf{83.6} & \textbf{88.2} & \textbf{82.2} & \textbf{84.4} & \textbf{100.0} & \textbf{100.0} \\
\addlinespace[1pt]
& \multirow{6}{*}{PDrop} & \multirow{2}{*}{40} & {\color{freegray}Free} & 55.6 & 78.8 & 38.0 & 83.8 & 74.9 & 83.9 & 80.3 & 77.6 & 43.5 & 50.1 & 43.3 & 48.3 & 92.5 & 79.4 \\
& & & {\color{postblue}Post} & 61.5 & 80.5 & 57.4 & 86.2 & 79.2 & 88.3 & 80.8 & 90.3 & 70.4 & 80.7 & 71.7 & 76.2 & 103.9 & 97.5 \\
& & \multirow{2}{*}{20} & {\color{freegray}Free} & 51.1 & 73.3 & 33.9 & 80.5 & 71.6 & 82.3 & 76.6 & 54.9 & 29.4 & 23.2 & 19.2 & 20.8 & 87.0 & 65.1 \\
& & & {\color{postblue}Post} & 57.9 & 76.6 & 54.2 & 85.0 & 76.6 & 85.8 & 79.3 & 72.7 & 53.9 & 65.5 & 56.2 & 60.9 & 99.7 & 87.5 \\
& & \multirow{2}{*}{10} & {\color{freegray}Free} & 39.3 & 52.4 & 15.5 & 76.5 & 65.5 & 62.6 & 55.3 & 19.9 & 13.1 & 8.4 & 5.8 & 6.9 & 66.8 & 44.1 \\
& & & {\color{postblue}Post} & 50.9 & 65.2 & 46.9 & 80.5 & 71.8 & 78.9 & 64.5 & 40.2 & 32.4 & 29.3 & 18.0 & 20.1 & 89.6 & 64.7 \\
\addlinespace[2pt]
& \multirow{6}{*}{PACT} & \multirow{2}{*}{40} & {\color{freegray}Free} & 60.3 & 81.6 & 41.8 & 86.9 & 81.7 & 87.2 & 80.7 & 85.7 & 75.3 & 79.8 & 73.3 & 75.9 & 98.3 & 94.8 \\
& & & {\color{postblue}Post} & 62.8 & 81.1 & 56.3 & 87.2 & 81.7 & 88.6 & 80.1 & 90.2 & 76.3 & 84.0 & 77.1 & 80.0 & 104.8 & 99.7 \\
& & \multirow{2}{*}{20} & {\color{freegray}Free} & 56.9 & 75.6 & 33.8 & 83.3 & 73.4 & 83.0 & 65.8 & 52.2 & 44.8 & 37.8 & 33.3 & 35.5 & 90.0 & 71.1 \\
& & & {\color{postblue}Post} & 60.7 & 78.3 & 54.9 & 84.2 & 78.2 & 87.1 & 74.7 & 69.9 & 59.5 & 66.5 & 58.4 & 60.9 & 101.5 & 88.6 \\
& & \multirow{2}{*}{10} & {\color{freegray}Free} & 51.6 & 68.9 & 26.1 & 80.2 & 68.4 & 77.5 & 53.6 & 30.7 & 32.9 & 14.1 & 12.7 & 12.6 & 81.6 & 56.1 \\
& & & {\color{postblue}Post} & 58.0 & 74.4 & 52.7 & 82.4 & 73.8 & 86.0 & 64.8 & 49.3 & 46.2 & 47.9 & 39.4 & 43.2 & 97.9 & 77.2 \\
\midrule
\multirow{13}{*}{\textbf{LLaVA-OV-1.5}} & \textbf{Base} & - & - & \textbf{62.2} & \textbf{81.9} & \textbf{48.7} & \textbf{98.5} & \textbf{84.4} & \textbf{89.2} & \textbf{79.6} & \textbf{97.0} & \textbf{81.7} & \textbf{56.3} & \textbf{50.5} & \textbf{51.1} & \textbf{100.0} & \textbf{100.0} \\
\addlinespace[1pt]
& \multirow{6}{*}{PDrop} & \multirow{2}{*}{40} & {\color{freegray}Free} & 56.0 & 75.1 & 39.4 & 90.8 & 76.5 & 85.0 & 66.2 & 58.2 & 26.4 & 24.6 & 20.4 & 19.9 & 90.2 & 70.0 \\
& & & {\color{postblue}Post} & 59.7 & 79.4 & 52.9 & 95.9 & 79.1 & 87.4 & 74.7 & 85.0 & 58.9 & 74.1 & 65.0 & 69.9 & 98.4 & 103.4 \\
& & \multirow{2}{*}{20} & {\color{freegray}Free} & 50.4 & 68.3 & 31.2 & 85.4 & 71.2 & 78.0 & 64.1 & 42.5 & 15.1 & 2.2 & 1.6 & 2.0 & 81.2 & 53.4 \\
& & & {\color{postblue}Post} & 53.6 & 71.2 & 49.7 & 87.8 & 73.1 & 81.4 & 66.9 & 58.2 & 42.9 & 56.9 & 45.6 & 50.5 & 90.3 & 85.7 \\
& & \multirow{2}{*}{10} & {\color{freegray}Free} & 44.9 & 55.1 & 15.1 & 78.3 & 68.7 & 69.8 & 37.0 & 19.7 & 6.1 & 0.4 & 0.3 & 0.4 & 68.3 & 40.5 \\
& & & {\color{postblue}Post} & 50.0 & 64.7 & 46.9 & 82.4 & 70.9 & 80.4 & 42.9 & 32.4 & 19.6 & 37.9 & 26.6 & 30.3 & 85.5 & 67.0 \\
\addlinespace[2pt]
& \multirow{6}{*}{PACT} & \multirow{2}{*}{40} & {\color{freegray}Free} & 61.0 & 80.4 & 46.1 & 97.6 & 82.5 & 88.6 & 77.1 & 87.5 & 69.7 & 21.3 & 18.5 & 19.3 & 97.9 & 81.0 \\
& & & {\color{postblue}Post} & 61.3 & 80.0 & 56.2 & 97.6 & 82.4 & 88.3 & 75.4 & 88.9 & 70.7 & 74.2 & 65.0 & 71.7 & 101.2 & 106.8 \\
& & \multirow{2}{*}{20} & {\color{freegray}Free} & 58.0 & 75.3 & 40.3 & 90.2 & 76.3 & 86.6 & 67.8 & 54.1 & 43.5 & 0.6 & 0.4 & 0.7 & 91.2 & 62.0 \\
& & & {\color{postblue}Post} & 58.8 & 77.8 & 53.6 & 92.7 & 78.3 & 87.8 & 70.6 & 66.7 & 51.1 & 53.0 & 42.0 & 49.2 & 97.5 & 89.9 \\
& & \multirow{2}{*}{10} & {\color{freegray}Free} & 52.8 & 69.1 & 34.8 & 84.8 & 72.3 & 82.3 & 54.6 & 26.9 & 32.8 & 0.2 & 0.1 & 0.1 & 84.2 & 53.5 \\
& & & {\color{postblue}Post} & 56.7 & 74.0 & 50.4 & 85.5 & 74.6 & 86.4 & 61.6 & 42.5 & 38.4 & 40.6 & 28.4 & 33.4 & 92.8 & 76.6 \\
\bottomrule
\end{tabular}
}
\caption{
Full LLM-level pruning results on Qwen2.5-VL-7B and LLaVA-OV-1.5-8B.
Ratio is the layer-averaged visual-token keep ratio in percent, and Free and Post denote the model before and after post-training.
}
\label{tab:pdrop_pact_free_post}
\end{table*}

\section{Grounding Details}

This section provides detailed grounding results on RefCOCO, RefCOCO+, and RefCOCOg. We report the key metrics on each split, including IoU, Acc@0.5, and Center Acc, to complement the aggregated grounding results in the main paper.

\begin{table*}[t]
\centering
\scriptsize
\setlength{\tabcolsep}{2.2pt}
\renewcommand{\arraystretch}{1.12}

\resizebox{\textwidth}{!}{
\begin{tabular}{lllccccccccccccccc}
\toprule
\multirow{3}{*}{\textbf{Level}} &
\multirow{3}{*}{\textbf{Method / Ratio}} &
\multirow{3}{*}{\textbf{Setting}} &
\multicolumn{6}{c}{\textbf{RefCOCO}} &
\multicolumn{6}{c}{\textbf{RefCOCO+}} &
\multicolumn{3}{c}{\textbf{RefCOCOg}} \\
\cmidrule(lr){4-9}
\cmidrule(lr){10-15}
\cmidrule(lr){16-18}
& & &
\multicolumn{3}{c}{testA} &
\multicolumn{3}{c}{testB} &
\multicolumn{3}{c}{testA} &
\multicolumn{3}{c}{testB} &
\multicolumn{3}{c}{test} \\
\cmidrule(lr){4-6}
\cmidrule(lr){7-9}
\cmidrule(lr){10-12}
\cmidrule(lr){13-15}
\cmidrule(lr){16-18}
& & &
IoU & Acc@0.5 & Center &
IoU & Acc@0.5 & Center &
IoU & Acc@0.5 & Center &
IoU & Acc@0.5 & Center &
IoU & Acc@0.5 & Center \\
\midrule
\multirow{12}{*}{Encoder} & \multirow{2}{*}{SAINT-40} & {\color{freegray}Free} & 40.6 & 39.9 & 68.6 & 40.6 & 39.8 & 71.4 & 38.8 & 38.2 & 66.0 & 36.5 & 35.6 & 64.9 & 39.5 & 38.2 & 68.1 \\
& & {\color{postblue}Post} & 68.7 & 82.0 & 93.9 & 61.3 & 72.5 & 88.9 & 62.0 & 73.2 & 87.9 & 52.0 & 59.9 & 77.5 & 61.0 & 72.0 & 88.0 \\
& \multirow{2}{*}{SAINT-20} & {\color{freegray}Free} & 19.4 & 10.9 & 41.0 & 21.9 & 11.6 & 48.0 & 18.4 & 10.3 & 39.6 & 20.2 & 10.9 & 44.7 & 19.3 & 10.1 & 41.3 \\
& & {\color{postblue}Post} & 64.3 & 77.9 & 91.2 & 57.3 & 67.3 & 87.9 & 58.8 & 70.3 & 85.2 & 49.2 & 56.2 & 75.8 & 56.8 & 67.1 & 85.9 \\
& \multirow{2}{*}{SAINT-10} & {\color{freegray}Free} & 11.4 & 3.8 & 29.9 & 12.3 & 4.3 & 30.3 & 10.9 & 3.7 & 28.4 & 11.7 & 3.7 & 28.9 & 10.4 & 2.6 & 26.6 \\
& & {\color{postblue}Post} & 55.7 & 65.2 & 85.8 & 49.3 & 55.2 & 82.9 & 49.1 & 56.4 & 77.6 & 41.3 & 44.3 & 71.3 & 48.5 & 54.7 & 80.4 \\
& \multirow{2}{*}{ToMe-40} & {\color{freegray}Free} & 40.0 & 40.7 & 67.4 & 40.0 & 38.5 & 71.7 & 38.5 & 38.6 & 65.5 & 35.9 & 33.9 & 63.8 & 38.8 & 37.3 & 68.1 \\
& & {\color{postblue}Post} & 68.9 & 82.6 & 93.1 & 62.2 & 74.9 & 89.1 & 64.2 & 76.9 & 87.3 & 54.1 & 63.5 & 78.2 & 62.2 & 74.2 & 87.5 \\
& \multirow{2}{*}{ToMe-20} & {\color{freegray}Free} & 21.1 & 13.4 & 44.3 & 23.0 & 13.3 & 49.4 & 20.4 & 12.6 & 43.6 & 20.9 & 12.3 & 44.9 & 20.8 & 11.4 & 44.2 \\
& & {\color{postblue}Post} & 63.4 & 77.3 & 91.8 & 56.8 & 67.6 & 87.5 & 57.9 & 70.2 & 85.1 & 48.9 & 56.6 & 75.6 & 56.5 & 67.3 & 85.2 \\
& \multirow{2}{*}{ToMe-10} & {\color{freegray}Free} & 12.3 & 4.4 & 31.0 & 12.9 & 4.3 & 30.9 & 11.7 & 4.2 & 30.4 & 12.3 & 4.0 & 29.5 & 11.3 & 3.1 & 28.5 \\
& & {\color{postblue}Post} & 56.0 & 65.1 & 85.8 & 50.1 & 55.6 & 83.5 & 49.2 & 56.3 & 77.7 & 41.3 & 43.3 & 70.1 & 48.6 & 53.9 & 80.0 \\
\midrule
\multirow{12}{*}{Projector} & \multirow{2}{*}{DivPrune-40} & {\color{freegray}Free} & 72.7 & 86.9 & 93.8 & 64.6 & 75.4 & 89.4 & 69.4 & 82.5 & 90.2 & 58.1 & 67.3 & 80.8 & 66.1 & 77.1 & 88.4 \\
& & {\color{postblue}Post} & 76.4 & 89.5 & 95.4 & 68.8 & 80.7 & 91.4 & 72.2 & 84.1 & 91.0 & 61.2 & 70.5 & 81.7 & 69.9 & 81.4 & 90.8 \\
& \multirow{2}{*}{DivPrune-20} & {\color{freegray}Free} & 55.6 & 65.9 & 88.7 & 47.2 & 50.3 & 80.0 & 51.7 & 61.0 & 82.9 & 41.0 & 42.7 & 71.0 & 48.2 & 52.6 & 79.7 \\
& & {\color{postblue}Post} & 71.2 & 85.4 & 93.7 & 63.8 & 76.1 & 89.2 & 66.5 & 79.5 & 88.2 & 56.0 & 65.5 & 78.8 & 64.6 & 76.8 & 88.5 \\
& \multirow{2}{*}{DivPrune-10} & {\color{freegray}Free} & 36.5 & 29.4 & 73.1 & 30.3 & 21.1 & 59.6 & 33.1 & 26.7 & 66.4 & 25.6 & 17.2 & 51.0 & 29.7 & 22.1 & 58.5 \\
& & {\color{postblue}Post} & 63.3 & 77.2 & 89.7 & 55.9 & 65.6 & 85.7 & 57.3 & 69.9 & 82.0 & 47.9 & 54.5 & 73.9 & 56.2 & 66.5 & 84.3 \\
& \multirow{2}{*}{AIM-40} & {\color{freegray}Free} & 71.1 & 86.8 & 94.7 & 63.4 & 76.7 & 90.9 & 68.2 & 83.1 & 91.3 & 57.5 & 68.5 & 82.7 & 63.8 & 76.6 & 89.5 \\
& & {\color{postblue}Post} & 76.2 & 90.1 & 95.8 & 68.6 & 81.5 & 91.8 & 71.5 & 84.0 & 91.3 & 60.7 & 70.2 & 82.0 & 69.4 & 82.1 & 91.3 \\
& \multirow{2}{*}{AIM-20} & {\color{freegray}Free} & 51.6 & 58.0 & 89.3 & 44.5 & 44.2 & 79.4 & 48.4 & 54.2 & 84.4 & 39.0 & 36.6 & 70.6 & 44.3 & 44.7 & 79.3 \\
& & {\color{postblue}Post} & 70.5 & 86.4 & 94.4 & 63.2 & 77.1 & 90.4 & 65.9 & 80.0 & 88.9 & 56.0 & 66.7 & 79.8 & 63.5 & 76.8 & 89.7 \\
& \multirow{2}{*}{AIM-10} & {\color{freegray}Free} & 32.2 & 19.6 & 73.7 & 27.0 & 15.7 & 57.3 & 29.5 & 17.8 & 68.6 & 23.4 & 12.6 & 49.7 & 25.9 & 14.6 & 57.5 \\
& & {\color{postblue}Post} & 64.0 & 78.6 & 91.7 & 57.2 & 67.7 & 88.7 & 59.0 & 71.7 & 85.3 & 49.5 & 56.9 & 76.9 & 56.9 & 67.6 & 86.3 \\
\midrule
\multirow{12}{*}{LLM} & \multirow{2}{*}{PDrop-40} & {\color{freegray}Free} & 48.9 & 54.8 & 76.4 & 43.6 & 45.4 & 72.9 & 43.9 & 49.1 & 68.5 & 37.1 & 37.6 & 60.5 & 44.8 & 48.3 & 71.5 \\
& & {\color{postblue}Post} & 71.3 & 85.5 & 92.7 & 64.1 & 75.9 & 88.3 & 66.1 & 78.3 & 86.6 & 55.7 & 65.2 & 77.4 & 65.1 & 76.2 & 87.7 \\
& \multirow{2}{*}{PDrop-20} & {\color{freegray}Free} & 31.1 & 26.5 & 61.4 & 28.4 & 19.9 & 58.4 & 27.2 & 22.4 & 53.4 & 23.8 & 16.0 & 47.6 & 27.5 & 20.8 & 55.4 \\
& & {\color{postblue}Post} & 58.1 & 69.9 & 85.3 & 53.1 & 61.1 & 82.0 & 50.9 & 61.0 & 75.0 & 45.4 & 51.4 & 70.4 & 51.7 & 60.9 & 79.7 \\
& \multirow{2}{*}{PDrop-10} & {\color{freegray}Free} & 18.8 & 7.7 & 41.6 & 18.7 & 9.1 & 41.6 & 16.3 & 5.8 & 32.0 & 15.9 & 5.8 & 30.1 & 15.2 & 6.9 & 29.4 \\
& & {\color{postblue}Post} & 36.4 & 31.8 & 64.4 & 34.6 & 26.8 & 64.9 & 26.5 & 21.4 & 46.4 & 23.9 & 14.6 & 44.9 & 27.0 & 20.1 & 49.5 \\
& \multirow{2}{*}{PACT-40} & {\color{freegray}Free} & 70.1 & 85.7 & 94.7 & 62.8 & 73.9 & 88.9 & 67.0 & 81.3 & 90.8 & 56.6 & 65.4 & 80.6 & 64.0 & 75.9 & 88.9 \\
& & {\color{postblue}Post} & 73.3 & 89.4 & 95.9 & 65.5 & 78.6 & 91.7 & 69.5 & 84.2 & 92.2 & 58.9 & 69.9 & 83.0 & 66.4 & 80.0 & 91.2 \\
& \multirow{2}{*}{PACT-20} & {\color{freegray}Free} & 43.6 & 43.2 & 79.9 & 38.3 & 32.4 & 73.9 & 40.2 & 38.9 & 74.2 & 33.6 & 27.7 & 64.0 & 39.5 & 35.5 & 73.8 \\
& & {\color{postblue}Post} & 60.4 & 73.5 & 90.6 & 52.5 & 59.5 & 86.8 & 55.6 & 67.0 & 83.9 & 45.6 & 49.8 & 75.0 & 53.1 & 60.9 & 84.4 \\
& \multirow{2}{*}{PACT-10} & {\color{freegray}Free} & 25.7 & 16.6 & 55.4 & 21.6 & 11.6 & 48.9 & 23.8 & 14.9 & 50.9 & 19.9 & 10.5 & 42.8 & 22.9 & 12.6 & 48.5 \\
& & {\color{postblue}Post} & 49.5 & 53.5 & 80.9 & 43.8 & 42.2 & 77.4 & 42.6 & 45.1 & 69.3 & 36.3 & 33.8 & 64.0 & 42.6 & 43.1 & 72.9 \\
\bottomrule
\end{tabular}
}

\caption{
Detailed grounding performance of Qwen2.5-VL-7B under encoder-, projector-, and LLM-level visual token pruning.
For each split, we report IoU, Acc@0.5, and Center Acc.
``Free'' denotes the training-free pruning result, while ``Post'' denotes the result after post-training.
All values are reported as percentages.
}
\label{tab:qwen_grounding_detail}
\end{table*}

\begin{table*}[htbp]
\centering
\scriptsize
\setlength{\tabcolsep}{2.2pt}
\renewcommand{\arraystretch}{1.12}

\resizebox{\textwidth}{!}{
\begin{tabular}{lllccccccccccccccc}
\toprule
\multirow{3}{*}{\textbf{Level}} &
\multirow{3}{*}{\textbf{Method / Ratio}} &
\multirow{3}{*}{\textbf{Setting}} &
\multicolumn{6}{c}{\textbf{RefCOCO}} &
\multicolumn{6}{c}{\textbf{RefCOCO+}} &
\multicolumn{3}{c}{\textbf{RefCOCOg}} \\
\cmidrule(lr){4-9}
\cmidrule(lr){10-15}
\cmidrule(lr){16-18}
& & &
\multicolumn{3}{c}{testA} &
\multicolumn{3}{c}{testB} &
\multicolumn{3}{c}{testA} &
\multicolumn{3}{c}{testB} &
\multicolumn{3}{c}{test} \\
\cmidrule(lr){4-6}
\cmidrule(lr){7-9}
\cmidrule(lr){10-12}
\cmidrule(lr){13-15}
\cmidrule(lr){16-18}
& & &
IoU & Acc@0.5 & Center &
IoU & Acc@0.5 & Center &
IoU & Acc@0.5 & Center &
IoU & Acc@0.5 & Center &
IoU & Acc@0.5 & Center \\
\midrule
Base & Base & - & 51.6 & 62.5 & 94.6 & 46.6 & 50.3 & 90.2 & 48.5 & 58.0 & 91.4 & 41.7 & 43.0 & 82.5 & 46.4 & 51.1 & 89.0 \\
\midrule
\multirow{12}{*}{Encoder} & \multirow{2}{*}{SAINT-40} & {\color{freegray}Free} & 43.1 & 38.0 & 90.2 & 39.1 & 31.8 & 84.3 & 39.8 & 34.5 & 85.6 & 34.3 & 26.7 & 75.7 & 38.8 & 32.1 & 82.3 \\
& & {\color{postblue}Post} & 68.1 & 85.2 & 94.8 & 60.4 & 74.0 & 91.3 & 64.0 & 79.3 & 91.1 & 53.8 & 65.0 & 82.1 & 61.6 & 75.7 & 90.8 \\
& \multirow{2}{*}{SAINT-20} & {\color{freegray}Free} & 30.7 & 12.5 & 80.8 & 28.1 & 11.7 & 72.2 & 27.2 & 10.3 & 74.7 & 23.9 & 9.4 & 62.6 & 27.5 & 10.9 & 71.3 \\
& & {\color{postblue}Post} & 62.6 & 77.1 & 93.1 & 55.3 & 65.4 & 88.8 & 56.8 & 68.9 & 87.8 & 48.2 & 55.6 & 79.3 & 55.5 & 66.7 & 88.7 \\
& \multirow{2}{*}{SAINT-10} & {\color{freegray}Free} & 19.0 & 3.2 & 63.3 & 18.7 & 2.9 & 60.7 & 15.6 & 2.5 & 54.8 & 15.1 & 2.4 & 49.2 & 17.3 & 2.7 & 55.1 \\
& & {\color{postblue}Post} & 55.2 & 65.0 & 86.1 & 50.0 & 54.9 & 85.2 & 48.5 & 55.6 & 77.5 & 42.4 & 45.5 & 72.8 & 49.2 & 54.8 & 81.6 \\
& \multirow{2}{*}{ToMe-40} & {\color{freegray}Free} & 43.2 & 38.7 & 91.0 & 38.8 & 31.2 & 83.9 & 39.6 & 34.0 & 86.1 & 34.0 & 25.4 & 75.5 & 38.5 & 31.5 & 82.6 \\
& & {\color{postblue}Post} & 68.5 & 85.2 & 94.3 & 61.0 & 74.4 & 91.5 & 64.4 & 79.2 & 90.7 & 54.0 & 64.7 & 82.0 & 62.1 & 76.2 & 90.8 \\
& \multirow{2}{*}{ToMe-20} & {\color{freegray}Free} & 30.9 & 13.0 & 81.0 & 28.3 & 12.2 & 72.8 & 26.9 & 10.3 & 74.1 & 24.3 & 9.2 & 64.4 & 27.4 & 11.2 & 71.0 \\
& & {\color{postblue}Post} & 62.5 & 77.2 & 92.5 & 55.8 & 65.9 & 89.8 & 57.8 & 70.5 & 87.7 & 48.7 & 56.5 & 79.0 & 56.1 & 67.2 & 88.4 \\
& \multirow{2}{*}{ToMe-10} & {\color{freegray}Free} & 17.8 & 2.5 & 61.3 & 18.4 & 2.9 & 59.2 & 14.5 & 1.9 & 53.0 & 14.5 & 1.9 & 47.4 & 16.7 & 2.2 & 54.2 \\
& & {\color{postblue}Post} & 54.9 & 65.5 & 86.5 & 50.2 & 55.7 & 85.1 & 49.3 & 57.4 & 79.4 & 42.6 & 45.8 & 73.3 & 49.5 & 55.5 & 82.1 \\
\midrule
\multirow{12}{*}{Projector} & \multirow{2}{*}{DivPrune-40} & {\color{freegray}Free} & 36.7 & 16.1 & 90.2 & 33.3 & 17.7 & 81.7 & 33.9 & 13.9 & 86.2 & 29.4 & 14.6 & 73.8 & 32.7 & 15.2 & 80.8 \\
& & {\color{postblue}Post} & 66.7 & 83.2 & 94.3 & 59.3 & 72.3 & 91.3 & 62.2 & 76.8 & 89.9 & 52.9 & 63.3 & 82.0 & 60.3 & 74.4 & 90.8 \\
& \multirow{2}{*}{DivPrune-20} & {\color{freegray}Free} & 24.4 & 2.9 & 80.9 & 23.4 & 4.8 & 70.2 & 21.9 & 2.3 & 75.6 & 20.1 & 3.7 & 62.5 & 22.5 & 3.4 & 70.4 \\
& & {\color{postblue}Post} & 62.0 & 76.6 & 91.9 & 54.4 & 63.7 & 88.1 & 56.9 & 69.2 & 85.9 & 47.8 & 55.2 & 77.4 & 55.9 & 66.9 & 88.1 \\
& \multirow{2}{*}{DivPrune-10} & {\color{freegray}Free} & 14.6 & 0.3 & 69.4 & 15.5 & 1.1 & 60.9 & 12.4 & 0.4 & 62.2 & 12.9 & 0.9 & 51.5 & 13.9 & 0.7 & 57.9 \\
& & {\color{postblue}Post} & 50.8 & 58.0 & 86.2 & 45.6 & 47.9 & 82.4 & 42.9 & 45.5 & 77.5 & 37.3 & 37.5 & 70.1 & 45.2 & 47.6 & 81.2 \\
& \multirow{2}{*}{AIM-40} & {\color{freegray}Free} & 40.3 & 25.9 & 91.2 & 36.3 & 24.5 & 83.2 & 37.1 & 22.2 & 86.9 & 31.9 & 19.8 & 75.6 & 36.0 & 23.3 & 82.0 \\
& & {\color{postblue}Post} & 67.5 & 84.1 & 94.7 & 59.8 & 72.0 & 91.5 & 63.2 & 78.8 & 90.7 & 53.0 & 63.4 & 82.3 & 61.2 & 75.2 & 91.0 \\
& \multirow{2}{*}{AIM-20} & {\color{freegray}Free} & 29.4 & 7.4 & 82.7 & 27.4 & 9.5 & 73.0 & 26.3 & 5.9 & 77.0 & 23.9 & 7.1 & 65.2 & 26.4 & 7.6 & 72.7 \\
& & {\color{postblue}Post} & 61.8 & 76.4 & 91.2 & 55.2 & 64.9 & 89.0 & 56.7 & 68.9 & 85.2 & 48.3 & 55.1 & 78.5 & 56.1 & 67.0 & 88.1 \\
& \multirow{2}{*}{AIM-10} & {\color{freegray}Free} & 18.6 & 1.4 & 69.9 & 19.2 & 2.6 & 62.4 & 15.7 & 1.0 & 63.0 & 16.0 & 1.8 & 53.7 & 16.9 & 1.4 & 59.9 \\
& & {\color{postblue}Post} & 54.5 & 64.0 & 86.4 & 49.1 & 53.3 & 84.0 & 47.4 & 53.8 & 76.4 & 41.2 & 42.7 & 72.1 & 47.9 & 52.5 & 81.2 \\
\midrule
\multirow{12}{*}{LLM} & \multirow{2}{*}{PDrop-40} & {\color{freegray}Free} & 37.6 & 26.7 & 87.5 & 34.2 & 22.6 & 81.2 & 33.5 & 23.0 & 79.9 & 29.2 & 17.8 & 71.0 & 32.4 & 19.9 & 77.1 \\
& & {\color{postblue}Post} & 64.6 & 79.3 & 92.6 & 57.6 & 68.9 & 88.8 & 59.0 & 71.7 & 87.2 & 49.7 & 58.3 & 78.5 & 57.7 & 69.9 & 87.4 \\
& \multirow{2}{*}{PDrop-20} & {\color{freegray}Free} & 18.8 & 1.8 & 67.6 & 17.1 & 2.6 & 58.3 & 15.7 & 1.3 & 57.7 & 14.2 & 1.9 & 47.7 & 16.0 & 2.0 & 55.1 \\
& & {\color{postblue}Post} & 53.3 & 61.9 & 82.8 & 47.8 & 51.8 & 81.7 & 43.8 & 50.3 & 70.6 & 38.1 & 40.8 & 66.5 & 45.2 & 50.4 & 74.7 \\
& \multirow{2}{*}{PDrop-10} & {\color{freegray}Free} & 10.3 & 0.3 & 52.6 & 10.1 & 0.5 & 50.7 & 8.2 & 0.2 & 44.4 & 8.2 & 0.4 & 40.3 & 9.0 & 0.4 & 44.0 \\
& & {\color{postblue}Post} & 41.8 & 42.6 & 69.6 & 37.7 & 33.2 & 68.9 & 31.8 & 31.2 & 53.8 & 27.2 & 22.1 & 50.4 & 33.2 & 30.3 & 58.9 \\
& \multirow{2}{*}{PACT-40} & {\color{freegray}Free} & 37.5 & 22.5 & 87.2 & 34.4 & 20.1 & 82.8 & 34.5 & 19.9 & 82.4 & 30.5 & 17.1 & 75.3 & 34.2 & 19.3 & 81.7 \\
& & {\color{postblue}Post} & 64.2 & 79.4 & 94.3 & 57.5 & 68.9 & 91.1 & 58.5 & 71.4 & 89.7 & 50.6 & 58.7 & 81.8 & 58.5 & 71.7 & 90.6 \\
& \multirow{2}{*}{PACT-20} & {\color{freegray}Free} & 15.2 & 0.6 & 61.9 & 14.9 & 0.7 & 61.5 & 12.9 & 0.3 & 55.2 & 12.3 & 0.5 & 52.6 & 14.6 & 0.7 & 58.0 \\
& & {\color{postblue}Post} & 50.8 & 56.9 & 85.1 & 46.9 & 49.2 & 84.3 & 43.1 & 46.1 & 76.7 & 38.6 & 38.0 & 72.4 & 45.9 & 49.2 & 81.5 \\
& \multirow{2}{*}{PACT-10} & {\color{freegray}Free} & 9.1 & 0.2 & 52.6 & 9.9 & 0.3 & 57.5 & 7.3 & 0.1 & 43.8 & 7.3 & 0.2 & 44.3 & 8.6 & 0.1 & 49.0 \\
& & {\color{postblue}Post} & 42.6 & 42.5 & 76.2 & 41.7 & 38.6 & 78.8 & 33.1 & 30.5 & 61.8 & 31.5 & 26.2 & 60.7 & 37.1 & 33.4 & 69.8 \\
\bottomrule
\end{tabular}
}

\caption{
Detailed grounding performance of LLaVA-OV-1.5-8B under encoder-, projector-, and LLM-level visual token pruning.
For each split, we report IoU, Acc@0.5, and Center Acc.
``Free'' denotes the training-free pruning result, while ``Post'' denotes the result after post-training.
All values are reported as percentages.
}
\label{tab:ov_grounding_detail}
\end{table*}